\documentclass[11pt]{article}

\usepackage[final]{acl}

\usepackage{times}
\usepackage{latexsym}
\usepackage{amsmath} 
\usepackage{enumitem}
 \usepackage{booktabs} 
\usepackage{array}    
\usepackage{float}
\usepackage[T1]{fontenc}
\usepackage{placeins}
\usepackage[table,dvipsnames]{xcolor}

\newcommand{\pair}[2]{#1 & #2}

\newcommand{\inc}[1]{\textcolor{Green}{\textbf{+#1}}}

\usepackage[dvipsnames]{xcolor}

\definecolor{OursRow}{HTML}{EDEDED}

\definecolor{Exp}{HTML}{DDE8FF}

\usepackage[utf8]{inputenc}
\usepackage{amssymb}
\usepackage{subcaption}
\usepackage{graphicx} 
\usepackage{booktabs} 
\usepackage{multirow}   
\usepackage{amsmath}  
\usepackage{wrapfig}   
\usepackage{natbib}    
\usepackage{subcaption}
\usepackage{microtype}

\usepackage{colortbl}  
\usepackage{xcolor}    
\definecolor{mygray}{gray}{0.9} 
\usepackage{stfloats}
\usepackage[para]{footmisc}

\definecolor{codeblue}{RGB}{0, 82, 147}     
\definecolor{codegreen}{RGB}{0, 128, 0}    
\definecolor{codegray}{RGB}{100, 100, 100}  
\definecolor{codeorange}{RGB}{230, 145, 56} 
\definecolor{darkerblue}{rgb}{0,0.08,0.45} 
\definecolor{royalblue}{RGB}{65,105,225}
\definecolor{lightblue}{RGB}{221,235,247}
\definecolor{fig3blue}{RGB}{47, 122, 232}  
\definecolor{fig3red}{RGB}{213, 32, 52}
\definecolor{fig3green}{RGB}{0, 137, 72} 
\definecolor{fig3yellow}{RGB}{217, 161, 5}
\definecolor{gray94}{gray}{.94}
\definecolor{gray90}{gray}{.90}

\definecolor{darkgreen}{RGB}{34,139,34}

\newcommand{\gray}[1]{\textcolor{gray}{#1}}

\newcolumntype{g}{>{\columncolor{gray94}}c} 
\newcolumntype{b}{>{\columncolor{lightblue}}c} 
\newcommand{\brow}[1]{\rowcolor{lightblue}{#1}} 
\newcommand{\eg}{\textit{e.g.}}

\newcommand{\longtok}[1]{\textcolor{teal}{\textit{#1}}}

\usepackage{inconsolata}

\usepackage[most]{tcolorbox} 
\usepackage{fancyhdr}        
\usepackage{graphicx}
\usepackage{xcolor}
\usepackage{tikz}
\usetikzlibrary{calc}
\definecolor{damoblue}{HTML}{0064E0}
\definecolor{damofg}{HTML}{1C2B33}   
\definecolor{damobg}{HTML}{2A039B}   

\fancypagestyle{firstheader}{
    \fancyhf{} 
    \fancyhead[R]{%
        \includegraphics[height=0.8cm]{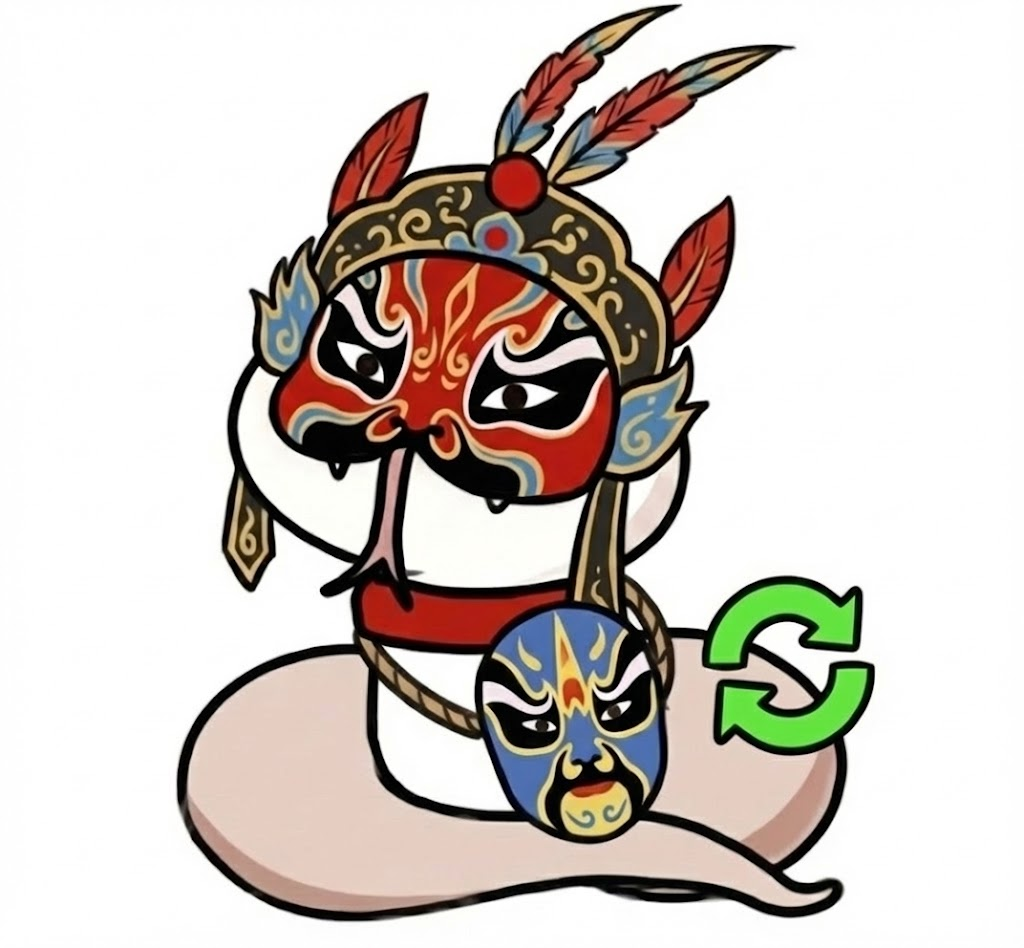} 
    }
}

\RequirePackage{fancyhdr}
\fancypagestyle{firstheader}{
	\fancyhf{}
	\fancyhead[R]{
		\includegraphics[height=0.8cm]{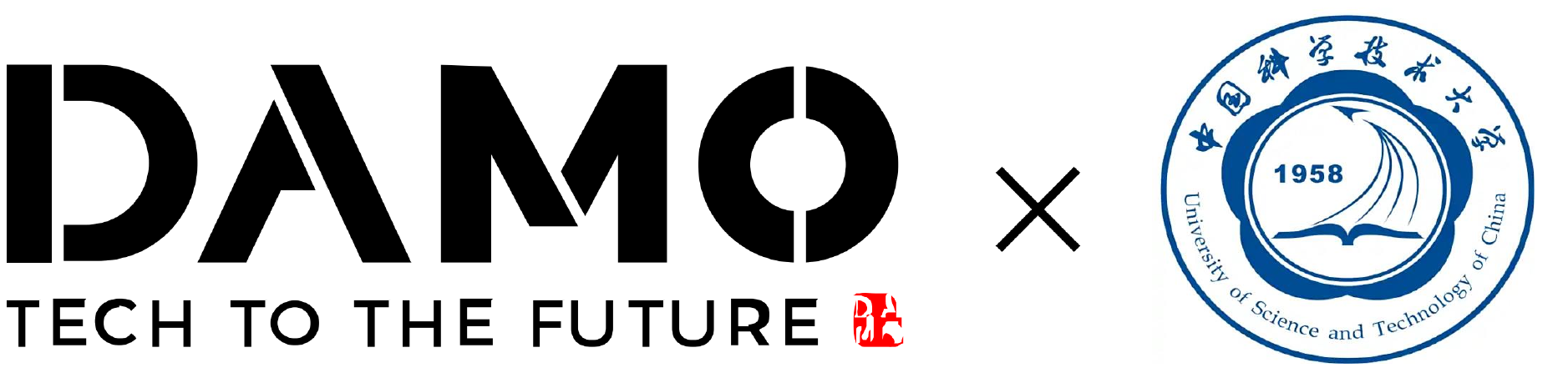}}
}

\newcommand{\CustomTitleBox}[3]{
    \begin{tcolorbox}[
        enhanced, 
        frame hidden,       
        colback=damobg!10,  
        arc=10pt,           
        boxrule=0pt,
        left=0.5cm, right=0.5cm, top=0.5cm, bottom=0.5cm, 
        overlay={
          \node[anchor=south east] at (interior.south east)
          [xshift=-8mm, yshift=8mm]
          {\includegraphics[width=1.5cm]{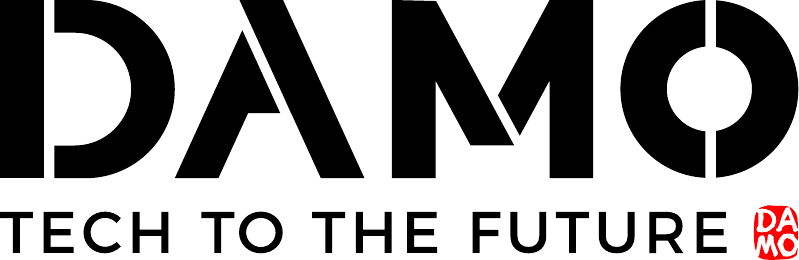}};
        },
    ]
        \setlength{\parindent}{0pt} 
        {\huge\bfseries\sffamily #1 \par} 
        \vspace{0.4cm}
        
        {\color{damobg} \large #2 \par} 
        \vspace{0.4cm}
        
        {\color{damofg} 
         \textbf{\sffamily Abstract} \par
         \vspace{0.1cm}
         #3
        } 
    \end{tcolorbox}
    \vspace{0.1cm} 
}

\begin{document}


\twocolumn[{
    \CustomTitleBox
    {
        \raisebox{-0.2\height}{\includegraphics[height=1.5em]{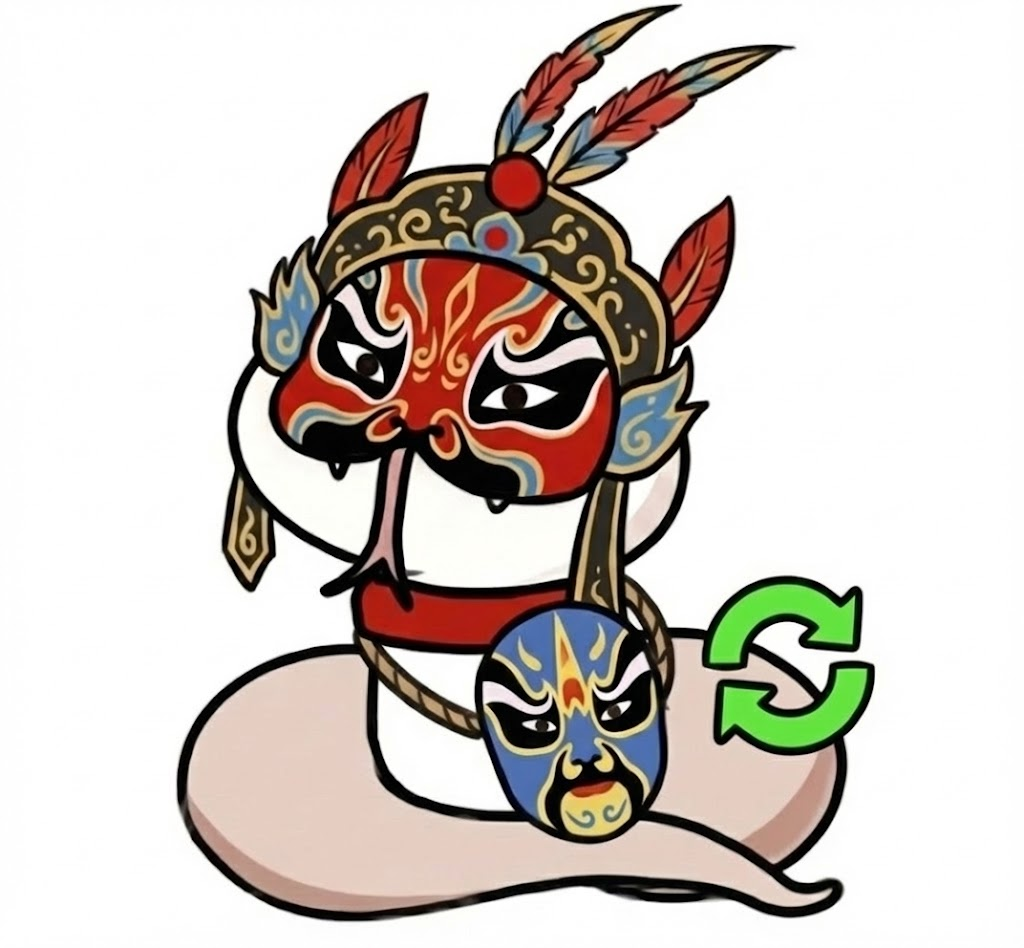}} 
        \space 
        I\textsuperscript{2}B-LPO: Latent Policy Optimization via Iterative Information Bottleneck
    }
    {   \normalsize
        \textbf{Huilin Deng}$^{1,2}$, \textbf{Hongchen Luo}$^{5}$, \textbf{Yue Zhu}$^{1,4}$, \textbf{Long Li}$^{1}$, 
         \textbf{Zhuoyue Chen}$^{1,3}$, \textbf{Xinghao Zhao}$^{1}$, \textbf{Ming Li}$^{2}$, 
         \textbf{Jihai Zhang}$^{1}$,\textbf{Mengchang Wang}$^{1}$,
         \textbf{Yang Cao}$^{2}$,\textbf{Yu Kang}$^{2}$
         \\
         
        \vspace{-5pt}
        \small
        $^1$DAMO Academy, Alibaba Group \quad $^2$University of Science and Technology of China \\
        $^3$ Zhejiang University \quad $^4$Shanghai Jiao Tong University \\
        $^5$ Northeastern University
    }
    {
       Despite recent advances in Reinforcement learning with verifiable rewards (RLVR) for large language model (LLM) reasoning, most methods suffer from exploration collapse, as the semantic homogeneity of random rollouts traps models in narrow, over-optimized behaviors. Existing methods leverage policy entropy to encourage exploration, but face inherent limitations: global entropy regularization is susceptible to reward hacking, inducing meaningless verbosity, whereas local token-selective updates struggle with the strong inductive bias of pre-trained models. To this end, we propose Latent Policy Optimization via Iterative Information Bottleneck (I\textsuperscript{2}B-LPO), which shifts from \textit{statistical perturbation of token distributions} to \textit{topological branching of reasoning trajectories}. I\textsuperscript{2}B-LPO triggers latent branching at high‑entropy states to diversify reasoning trajectories and applies the Information Bottleneck as a trajectory filter and self-reward to ensure concise and informative exploration. Empirical results on four mathematical benchmarks demonstrate that I\textsuperscript{2}B-LPO achieves state-of-the-art performance, with margins of up to 5.3\% in accuracy and 7.4\% in diversity metrics.

       \par \vspace{0.2cm} 
       
       {
           \small
           \setlength{\parskip}{0.2em} 
           
           {\sffamily\bfseries Date:} \today \par
           
           {\sffamily\bfseries Code:} \url{https://github.com/denghuilin-cyber/IIB-LPO} \par
           
           {\sffamily\bfseries Correspondence:} \href{mailto:forrest@ustc.edu.cn}{\texttt{forrest@ustc.edu.cn}} \par
       } 
    }
}]

\thispagestyle{firstheader}


\section{Introduction}


Reinforcement learning with verifiable rewards (RLVR) has emerged as a key method for LLM reasoning, particularly in tasks with deterministic verification such as mathematics \cite{deepseekR1,qwen25math}. This paradigm trains models to rollout and differentiate between multiple reasoning paths, reinforcing trajectories that lead to correct solutions while penalizing incorrect paths \cite{entropy_adv}. This contrastive approach has driven remarkable success across mathematical and broader reasoning tasks.



Despite this success, RLVR methods can unintentionally cause \textbf{exploration collapse} \cite{rethinkingEntropy}, where models converge on narrow, over-optimized behaviors and lose their incentive to explore alternative strategies. This pathology stems from the semantic homogeneity of random rollouts \cite{pathdiver}. While these rollouts exhibit variations in surface phrasing, the underlying reasoning rapidly degenerates into a few high-probability \textit{reasoning templates} \cite{modecollapse}, i.e., they share nearly identical reasoning patterns. Consequently, comparing such homogeneous paths yields vanishing advantage differentials and thus uninformative learning signals for policy optimization.


\begin{figure*}[htbp]
    \centering
    \includegraphics[width=0.95\linewidth]{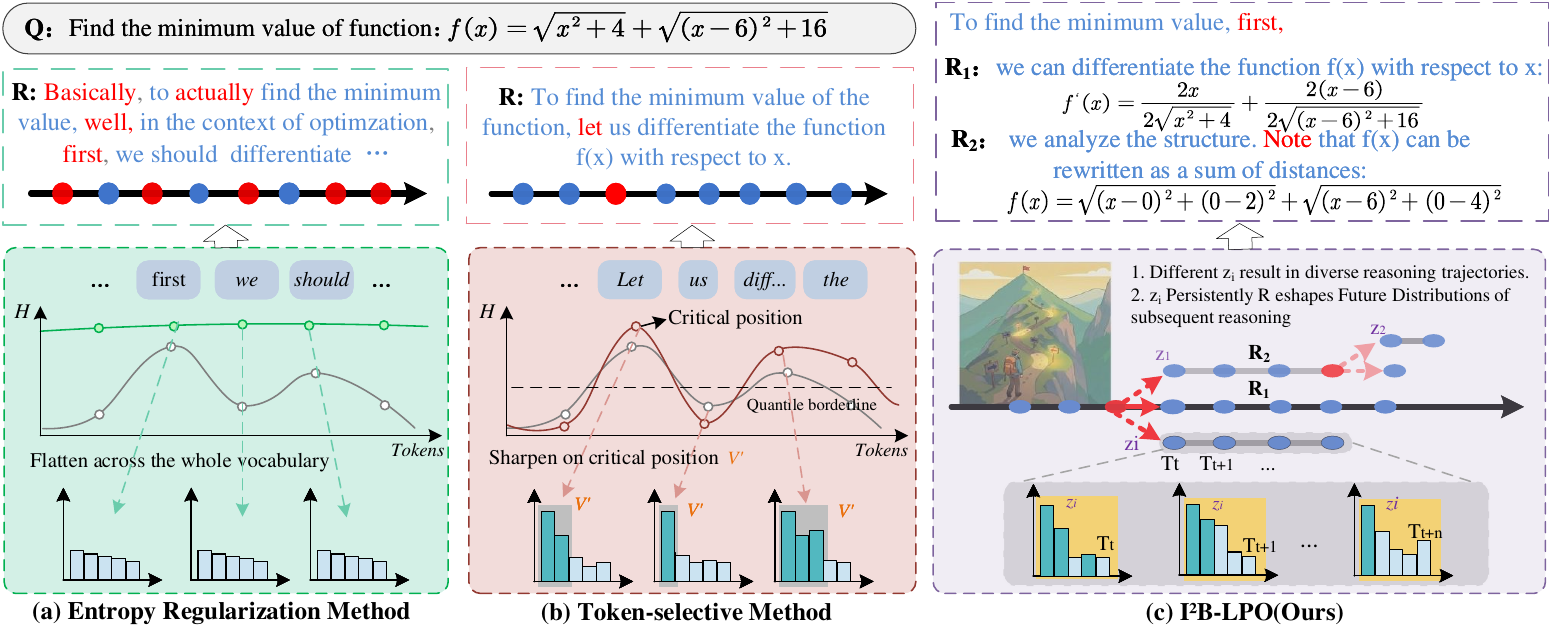}
    \vspace{-3pt}
    \caption{\textbf{Comparison of exploration paradigms in RLVR.} (a) Entropy Regularization globally smooths the probability distribution, leading to high-entropy yet meaningless verbosity. (b) Token-selective Methods locally sharpen the distribution; synonym replacement at these isolated points cannot overcome inductive biases. (c) I\textsuperscript{2}B-LPO introduces topological branching via latent variables $z$, resulting in distinct reasoning trajectories (e.g., Differentiation-based $R_1$ vs. Geometry-based $R_2$).}
    \label{fig1}
    \vspace{-8pt}
\end{figure*}

To mitigate this, policy entropy has been strategically leveraged to encourage exploration \cite{2_8_entropy}. Existing approaches typically fall into two paradigms. Entropy-regularization methods \cite{chao2024maxentropy} directly maximize the entropy of generated instances by globally smoothing the token-wise probability distribution (Fig. \ref{fig1}(a)). However, this global smoothing is susceptible to \textbf{reward hacking}---incentivizing the model to generate semantically vacuous verbosity \cite{yao2024tree}. As shown in Fig. \ref{fig1}(a), the model tends to generate excessive meta-discourse irrelevant to the problem. Conversely, token-selective methods \cite{covclip} amplify policy updates on high-entropy tokens, locally sharpening the next-token distribution at critical positions (Fig. \ref{fig1}(b)). Yet, such local sharpening struggles against the \textbf{inductive biases}. High-entropy positions frequently correspond to lexical ambiguity (e.g., synonym choice); merely sharpening the distribution at these points is insufficient to overcome the strong prior of pre-trained models \cite{casper2023open}. Consequently, both paradigms are confined to probabilistic perturbations, lacking the capacity to induce structural diversification in the reasoning process itself.




To address these limitations, this paper proposes a fundamental paradigm shift: moving from \textit{statistical perturbation} of token distributions to \textit{topological bifurcation of reasoning trajectories} (Fig. \ref{fig1}(c)). We introduce Latent Policy Optimization via Iterative Information Bottleneck (I\textsuperscript{2}B-LPO), which triggers latent branching at high-entropy states to shatter inductive biases and utilizes information-theoretic constraints to curb reward hacking. As illustrated in Fig. \ref{pipeline}, the method operates through two key mechanisms: (1) Entropy-driven Latent Branching utilizes a Conditional Variational Autoencoder (CVAE) \cite{CVAE_story_generation} to sample diverse latent variables $z_i$ at detected bifurcation (high-entropy states). Each $z_i$ serves as a structural prompt, injected into the LLM’s attention layers, continually steering the trajectory of subsequent reasoning. (2) Information Bottleneck Regularization functions as a dual-purpose filter and self-reward. By quantifying the trade-off between rationale compression and predictive power \cite{IB}, the IB objective identifies compact, informative paths for policy updates while simultaneously penalizing semantically vacuous verbosity.


Empirical results demonstrate I\textsuperscript{2}B-LPO's SOTA performance in both reasoning accuracy and semantic diversity, without excessive generation length. The significant margins in accuracy (5.3\%) and diversity (7.4\%) validate its efficacy in prompting exploration. The main contributions are summarized as follows:
(1) Paradigm shifts from statistical token perturbation to topological trajectory branching.
(2) Entropy-Driven Latent Branching explicitly restructures the reasoning topology to induce trajectory diversity.
(3) Dual-Purpose IB uses it both as a trajectory filter and self-reward, favoring concise, informative reasoning.
(4) SOTA performance in both accuracy and diversity without excessive length.


\section{Related Work}

\paragraph{Entropy Regularization in RL}
Entropy, which quantifies the dispersion of vocabulary distribution, reflects the predictive uncertainty of LLMs. Thus, it's strategically utilized to guide the exploratory behavior of LLMs \cite{chao2024maxentropy}. 
Existing methods typically fall into two categories. Token-selective methods target high-entropy tokens for selective updates \cite{2_8_entropy}. Regularization-based methods, such as concurrent works by \cite{wang2025RL_for_entropy} and \cite{entropy_adv}, incorporate an entropy regularizer directly into the loss or advantage function. Most priors rely on indiscriminate regularizer on the token distribution. Instead, we propose a paradigm shift towards conditional branching for effective exploration.

\paragraph{Self-Rewarding Reasoning.}
Recent RL advancements in LLMs rely heavily on outcome-based rewards, yet such sparse supervision neglects intermediate reasoning validity. As step-level labels require substantial human effort, self-rewarding methods have gained increasing attention. \cite{bai2022} and \cite{yuan2025selfrewarding} use predefined rules to generate preference pairs for Direct Preference Optimization (DPO) training. \cite{franken2024self} maximize mutual information between principles and responses, while \cite{zhang2025} is based on semantic entropy clustering. SPINE \cite{wu2025spine} adopts a majority-vote-based reward for self-consistency. However, unlike SPINE, which suppresses high-entropy states via static voting, we actively leverage them as gateways for topological branching and iterative trajectory refinement.

\section{Preliminary Analysis}

\begin{figure}[!t]
    \centering
    \includegraphics[width=1.0\columnwidth]{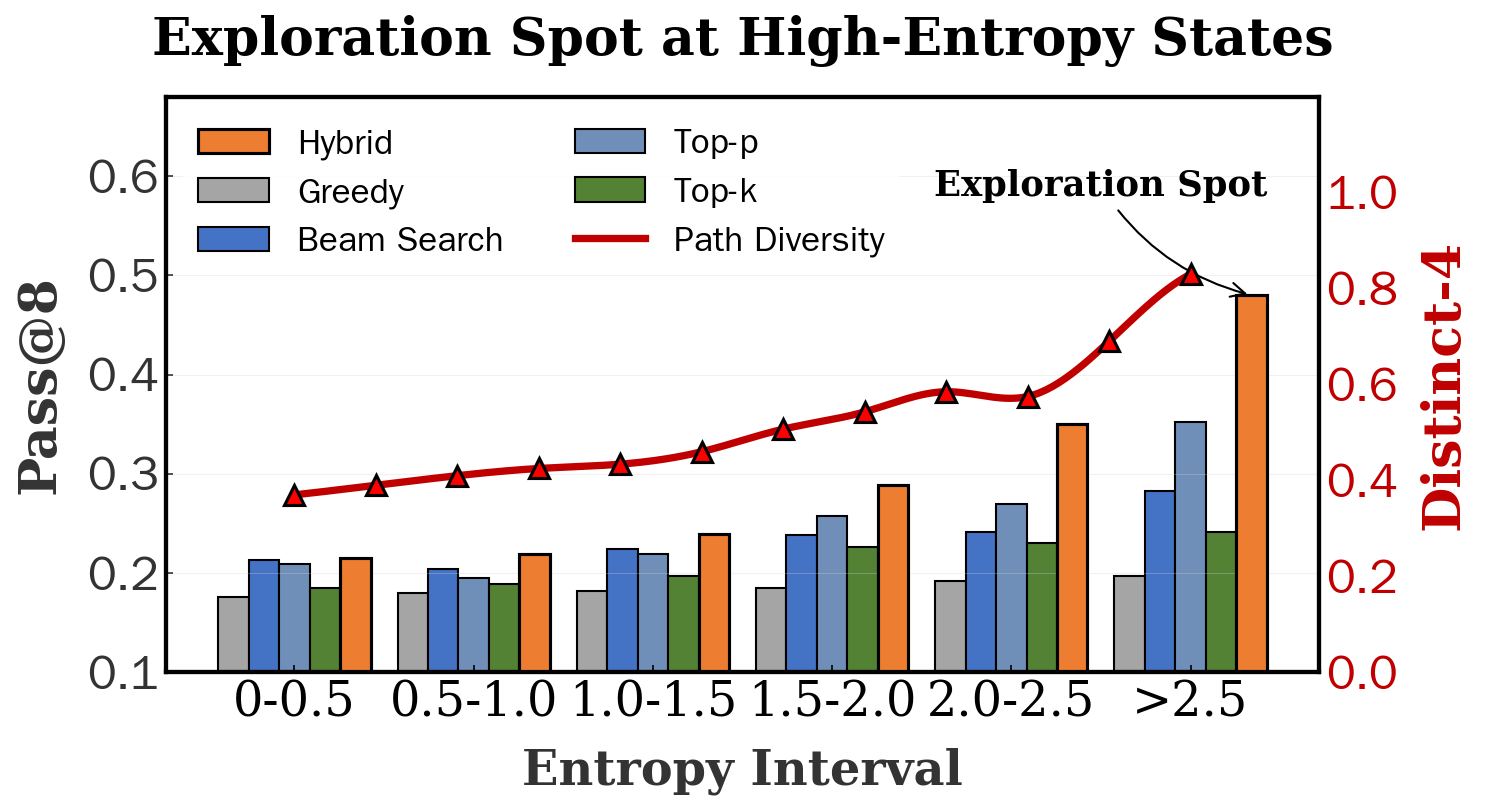}
    \vspace{-12pt}
   \caption{Performance of various decoding strategies trained on DeepMath. For each problem, we sample truncation points across entropy intervals to simulate varied exploration behaviors.}
    \label{fig:5_zhu_1_xian}
    \vspace{-10pt}
\end{figure}

\noindent\textbf{Pivotal Decision Points.} To validate whether entropy signals pivotal reasoning steps, we sampled prefixes conditioned on token-level entropy to simulate different exploration strategies via various decoding methods. Fig. \ref{fig:5_zhu_1_xian} shows that in high‑entropy intervals (>2.5) hybrid strategy substantially outperforms baselines—a performance gap far larger than in low‑entropy regions (<1.0). This confirms high-entropy states as natural decision points; branching here unlocks exploration potential.

\noindent\textbf{Verbosity without Gain.} As shown in Fig. \ref{fig:IB_pivot}, when applying standard GRPO, the accuracy plateaus early while the response length continues to rise, accompanied by a growing 4‑gram repetition rate (gray bars). This indicates that without additional guidance, the model tends to produce verbose and repetitive reasoning without genuine performance gains. Therefore, we introduce the IB self‑reward to suppress such unproductive expansion.

\section{Methodology}

\subsection{Problem Formulation}


Formally, let \textbf{q} denote input prompt, \textbf{r} the reasoning trajectory. We initialize the policy optimization with GRPO, starting with $M$ initial rollouts $\mathcal{R}_o=\{r_1, \dots, r_M\}$. I$^2$B-LPO comprises two phases:
\begin{itemize}[leftmargin=*, noitemsep, topsep=0pt, parsep=0pt]
    \item \textbf{Entropy-driven Branching}: For each base rollout $r_i$, we generate $K$ distinct branches while retaining the original one.
    This yields a branching set $\mathcal{R} = \{r_{i,j} \mid 1 \le i \le M, 1 \le j \le K+1\}$, containing $M \times (K+1)$ paths. Each path $r_{i,j}$ consists of a sequence of tokens $(o_1, \dots, o_{T_{i,j}})$. 
    \item \textbf{IB-Pruning}: we eliminate samples with low IB-scores from $\mathcal{R}$, retaining a high-quality subset $\mathcal{R^*}\in R$ such that $|\mathcal{R^*}| = N$.
\end{itemize}




\subsection{Entropy-driven Latent Branching}\label{sec:branching}


\subsubsection{Entropy-Driven Bifurcation Detection}\label{sec:branching_point}



For a given trajectory $r=(o_1, \dots, o_T)$, we identify a ``bifurcation point'' $t^* \in \{1, \dots, T\}$ corresponding to a state of high uncertainty. The token-level entropy $H_t$ at step $t$ is computed as:

\begin{figure}[!t]
    \centering
    \includegraphics[width=0.93\columnwidth]{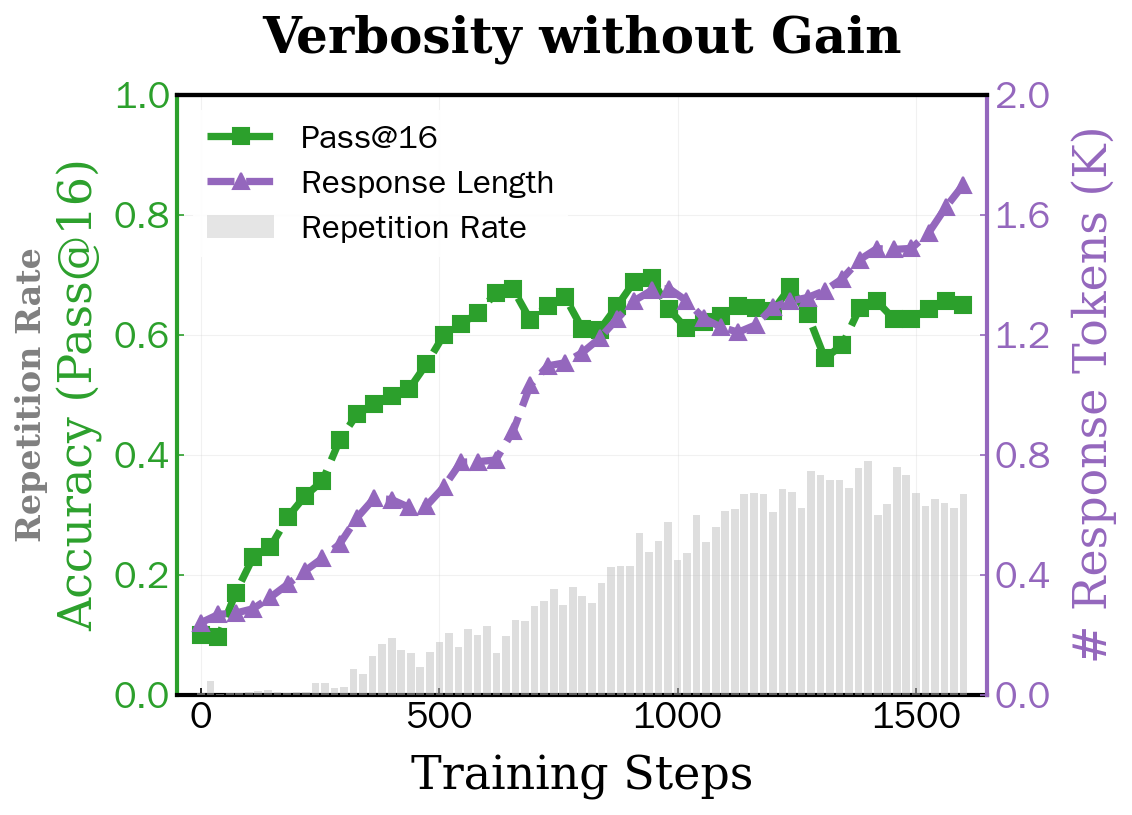}
    \caption{Accuracy and Response Length under GRPO. Notably, gray bars denote 4-gram repetition rate.}
    \vspace{-10pt}
    \label{fig:IB_pivot}
\end{figure}

\begin{figure*}[tbp]
    \centering
    \includegraphics[width=0.975\linewidth]{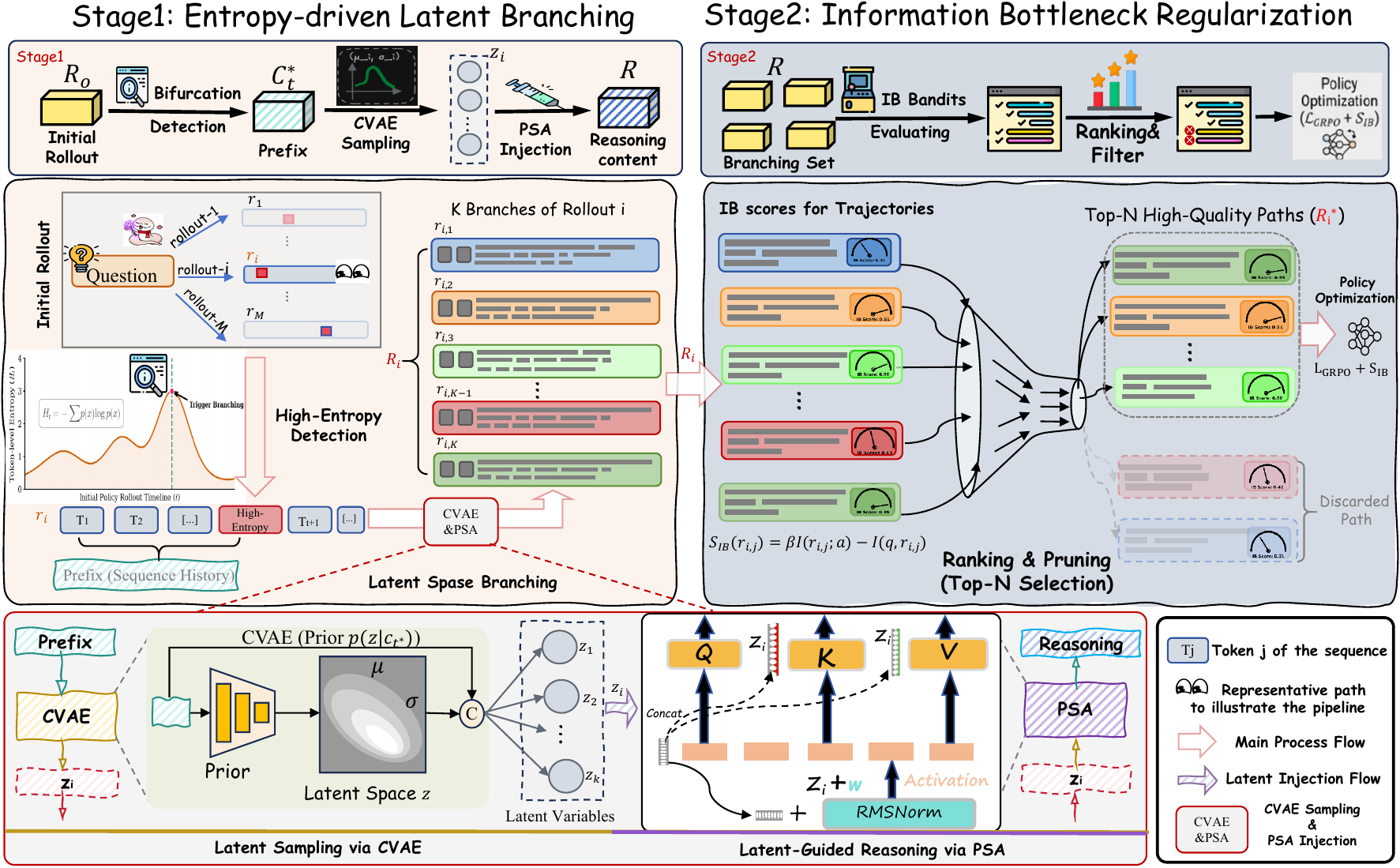}
    \caption{\textbf{Pipeline of the I\textsuperscript{2}B-LPO.} We use a representative path $r_i$ from the initial set $R_o$ to illustrate the workflow, which operates in two phases. (1) \textbf{Entropy-driven Latent Branching} expands $r_i$ into the branching set $R_i$ via Latent Sampling and PSA Injection, which are depicted in the bottom section.
(2) \textbf{Information Bottleneck Regularization} applies IB as a dual-purpose filter and self-reward to ensure concise and informative exploration. }
    \label{pipeline}
\end{figure*}

{\small
\begin{equation}
    H_t = -\sum_{v \in \mathcal{V}} P(v \mid q, o_{<t}) \log P(v \mid q, o_{<t}),
\end{equation}}
where $\mathcal{V}$ is the vocabulary. Following, we select the top 5\% highest‑entropy steps as candidate $\Omega$ ($\tau$: the 95$^{\text{th}}$ percentile of entropy history $\mathcal{H}$):
\begin{equation}
    \Omega = \{t \mid H_t \ge \tau\}, \quad t^* \sim \text{Uniform}(\Omega).
\end{equation}
Finally, we extract the prefix context $c_{t^*}$ by concatenating the query $q$ with the partial path preceding the split $c_{t^*} = [q, o_1, \dots, o_{t^*-1}]$.




\subsubsection{Latent Sampling via CVAE}\label{sec:latent_sampling}



Given the prefix $c_{t^*}$, we employ a separately trained \textbf{Conditional Variational Autoencoder (CVAE)} to sample latent variables $z$. Implementation of CVAE is detailed in Appendix~\ref{CVAE_details}. Formally, the CVAE models the conditional distribution of a solution trajectory $y$ given context via a latent variable: 

{\small
\begin{equation}
    p(y \mid x) = \int_z p(y \mid z, x) \, p(z \mid x) \, dz,
\end{equation}}
where the \textbf{prior network} $p(z \mid x)$ captures the distribution of $z_i$. During training, for each $r_i$, we sample $K$ independent latent codes from prior:

{\small
\begin{equation}
    z^{(j)} \sim p(z \mid c_{t^*}), \quad j = 1, \dots, K.
\end{equation}}


\subsubsection{Latent-Guided Reasoning via PSA}\label{PSA_guidance}




This section explores \textbf{\textit{how latent variables $\mathbf{z}_j$, generated by a CVAE, influence subsequent reasoning in LLMs.}} We employ \textbf{Pseudo Self-Attention (PSA)}, incorporating the latent code \( z \in \mathbb{R}^d \) into the self-attention mechanism at a per-layer basis. The detailed injection is as follows:

\begin{itemize}[leftmargin=*, noitemsep, topsep=0pt, parsep=0pt]

\item \textit{Phase 1: Adaptive Norm Modulation.} 
The latent code is projected and added to the learnable scaling parameter of RMSNorm:

\begin{equation}
\small
    w'_{l} = w_l + \gamma(t) \cdot \text{Proj}_{\phi}(z_j),
\end{equation}
where $\gamma(t)$ is a dynamic decay factor that anneals the latent influence over time. 

\item \textit{Phase 2: Augmented Self-Attention.} 
The latent code is further projected into modulation vectors and concatenated with original keys and values:
\end{itemize}

\vspace{-10pt}
\begin{equation}
\small
    K'_j = \begin{bmatrix} z_jK \\ K \end{bmatrix}, \quad
    V'_j = \begin{bmatrix} z_jV \\ V \end{bmatrix} 
    \in \mathbb{R}^{(1+l) \times d}
\end{equation}
where the notation $(\cdot)$ indicates row-wise concatenation. The enhanced attention is computed as:
{\begin{equation}
\setlength{\abovedisplayskip}{3pt}
\setlength{\belowdisplayskip}{3pt}
\small
\text{PSA}(Q, K'_j, V'_j) = \text{softmax}\left(\frac{QK'^T_j}{\sqrt{d_k}}\right)V'_j.
\end{equation}}

Thus, $z_i$ implicitly shifts features via RMSNorm and explicitly guides attention as a structural prompt, thereby steering the reasoning trajectory.



\subsection{Information Bottleneck Regularization}
 

\subsubsection{IB Score: Theory and Computation}
\label{sec:ib_foundation}

This section formulates the IB score and its approximation. We formulate LLM reasoning as an optimization trade-off between \textbf{rationale compression} and its \textbf{predictive power}. Specifically, an optimal policy minimizes prompt retention while maximizing the informativeness of final answers:

{\small
\begin{equation}
\label{eq:general_ib}
\min \mathcal{L}_{\text{IB}}(r) = I(q; r) - \beta I(r; a),
\end{equation}}
where $I(\cdot;\cdot)$ denotes Mutual Information (MI). The MI between $r$ and $a$ can be decomposed as:

{\small
\begin{equation}
\label{eq:ib_entropy_decomp}
I(r; a) = {H(r)} - {H(r \mid a)},
\end{equation}}
where $H(\cdot)$ denotes entropy. Eq. \ref{eq:ib_entropy_decomp} implies that maximizing $I(r; a)$ balances exploration (diversity $H(r)$) and precision (low ambiguity $H(r|a)$).

Formally, we define $S_{\text{IB}}(r) \equiv -\mathcal{L}_{\text{IB}}(r)$ at the trajectory-level, where higher values indicate a better balance between compression and informativeness. Based on the derivation in Appendix.~\ref{IB_Derivation}, we approximate the \textbf{$S_{\text{IB}}(r)$} as:

\vspace{-1em}

{\small
\begin{equation}
\label{eq:ib_score_approx}
\begin{split}
     S_{\text{IB}}(r) 
    &= \frac{1}{T} \sum_{t=1}^{T} \Big( \log \pi(o_t \mid o_{<t}, q) + \lambda \cdot H(o_t \mid o_{<t}, q) \Big) \\
    &= \frac{1}{T} \sum_{t=1}^{T} \mathcal{A}_{t} H(o_t \mid o_{<t}, q),
\end{split}
\raisetag{15pt}  
\end{equation}}
where $T$ is the response length. $H(o_t \mid o_{<t}, q)$ and $\mathcal{A}_t$ denote the policy entropy and advantage function at step $t$, respectively.


\subsubsection{IB-Guided Pruning and Optimization}
\label{sec:ib_optimization}

Given the candidate set $\mathcal{R}$ derived from entropy-driven branching, we refine the pool of reasoning paths by retaining the top-$N$ trajectories based on their IB scores. Formally, obtaining this optimal subset $\mathcal{R}^*$ is equivalent to solving:

{\small
\begin{equation}
    \mathcal{R}^* = \operatorname*{arg\,max}_{\substack{\mathcal{S} \subseteq \mathcal{R} \\ |\mathcal{S}| = N}} \sum_{r \in \mathcal{S}} S_{\text{IB}}(r),
    \label{eq:subset_selection}
\end{equation}}
where $\mathcal{S}$ denotes any subset of $\mathcal{R}$ with $|\mathcal{S}| = N$. In practice, we first compute $S_{\text{IB}}(r)$ for each $r \in \mathcal{R}$ using Eq.~(\ref{eq:ib_score_approx}), sort the results, and retain only the top $N$ paths. The pruned set $\mathcal{R}^*$ then serves as the training data for policy optimization. We also incorporate the IB score as an \textbf{auxiliary maximization objective} during training:

{
\begin{equation}
\setlength{\abovedisplayskip}{3pt}
\setlength{\belowdisplayskip}{3pt}
\small
    \mathcal{S}_{\text{IB}}(\theta; \mathcal{R}^*) =  \frac{1}{N} \sum_{r \in \mathcal{R}^*} S_{\text{IB}}(r).
\end{equation}}
{\small
\begin{equation}
\mathcal{J} = \mathcal{J}_{\text{GRPO}} + \gamma \cdot \mathcal{S}_{\text{IB}}(\theta; \mathcal{R}^*) ,
\end{equation}}
where $\gamma$ is a coefficient balancing task correctness and reasoning efficiency.

\begin{table*}[t]
    \centering
    \caption{\textbf{Overall Performance Comparison across 4 Mathematical Benchmarks.} We evaluate I\textsuperscript{2}B-LPO on two different backbones: \textbf{Qwen2.5-7B} and \textbf{Qwen3-14B}. We report \textbf{Pass@n} accuracy (\%) alongside average response length (\textbf{\#Tok}). \textbf{Bold} denotes the highest accuracy, while \longtok{colored italics} indicate the longest average response length.}
    \vspace{-0.30em}
    \setlength\tabcolsep{0.6em} 
    \resizebox{1.0\linewidth}{!}{
    \renewcommand{\arraystretch}{0.8} 
        \begin{tabular}{l|cccccccccccc}
        \toprule
\bf{Method} & \multicolumn{3}{c}{\bf{AIME2025}} & \multicolumn{3}{c}{\bf{AIME2024}} & \multicolumn{3}{c}{\bf{MATH-500}} & \multicolumn{3}{c}{\bf{Olympiad}} \\
& \bf{P@1} & \bf{P@256} & \bf{\#Tok} & \bf{P@1} & \bf{P@256} & \bf{\#Tok} & \bf{P@1} & \bf{P@16} & \bf{\#Tok} & \bf{P@1} & \bf{P@16} & \bf{\#Tok} \\ \hline

\multicolumn{13}{l}{\gray{\textit{Qwen2.5-7B}}} \\
+GRPO (Standard) & 8.2 & 50.0 & 1292 & 10.3 & 46.7 & 776 & 54.4 & 58.4 & 661 & 44.9 & 61.6 & 762 \\
\rowcolor{gray!10}Entropy-Reg.    & 9.3 & 47.7 & \longtok{3198} & 12.7 & 55.6 & \longtok{2358} & 57.4 & 70.4 & \longtok{1802} & 48.9 & 63.4 & 1823 \\
\rowcolor{gray!10}Entropy-Adv     & 11.8 & 53.3 & 2187 & 13.6 & 56.7 & 1424 & 58.5 & 74.0 & 1223 & 51.8 & 64.6 & \longtok{1894} \\
\rowcolor{gray!10}KL-Cov        & 11.3 & 52.1 & 1878 & 15.2 & 72.3 & 1283 & 68.2 & 82.1 & 1488 & 53.0 & 65.4 & 1190 \\
\rowcolor{gray!10}80/20           & 10.2 & 47.9 & 1898 & 16.2 & 70.5  & 1315 & 61.8 & 79.1 & 1190 & 49.6 & 62.2 & 1479 \\
SPINE         & 11.2 & 52.7 & 1634 & 14.4 & 68.2 & 1179 & 76.2 & 86.5 & 1104 & 52.5 & 67.8 & 1389 \\
SRLM            & 9.7 & 48.6 & 1574 & 15.1 & 65.1& 1334 & 73.4 & 84.7 & 1058 & 46.5 & 61.0 & 1167 \\
\brow{I\textsuperscript{2}B-LPO (Ours)} & \textbf{13.6} & \textbf{55.0} & 1465 & \textbf{18.6} & \textbf{79.7} & 1245 & \textbf{81.5} & \textbf{90.5} & 1080 & \textbf{58.0} & \textbf{69.5} & 1172 \\
\hline

\multicolumn{13}{l}{\gray{\textit{Qwen3-14B}}} \\
+GRPO (Standard) & 27.0 & 62.3 & 2045 & 34.4 & 67.8 & 1437 & 89.2 & 92.9 & 825 & 55.7 & 67.8 & 894 \\
\rowcolor{gray!10}Entropy-Reg.    & 28.8 & 59.8 & \longtok{4285} & 37.3 & 58.9 & \longtok{2963} & 90.2 & 93.0 & \longtok{2387} & 58.7 & 68.6 & 1880 \\
\rowcolor{gray!10}Entropy-Adv & 27.8 & 61.3 & 2287 & 42.8 & 63.1 & 2265 & 89.5 & 93.5 & 1626 & 57.8 & 70.6 & 1709 \\
\rowcolor{gray!10}KL-Cov & 34.6 & 62.8 & 2390 & 45.4 & 80.6 & 2018 & 91.7  & 93.6 & 1798 & 62.0 & 76.8 & 1801 \\
\rowcolor{gray!10}80/20    & 33.5 & 66.2 & 2168 & 43.9 & 78.5 & 1989 & 91.6 & 94.1 & 1488 & 60.8 & 73.2 & \longtok{2290} \\
SPINE & 28.4  & 59.6 & 2034 & 39.2 & 70.6 & 1966 & 89.9 & 93.2 & 1392 & 65.5 & 78.3 & 1643 \\
SRLM  & 29.7 & 58.3 & 2201 & 37.2 & 69.4 & 1876 & 91.1 & 94.6 & 1892 & 62.5 & 71.7 & 1686 \\
\brow{I\textsuperscript{2}B-LPO (Ours)} & \textbf{38.3} & \textbf{68.1} & 2465 & \textbf{46.6} & \textbf{82.5} & 1830 & \textbf{93.5} & \textbf{95.7} & 1432 & \textbf{68.0} & \textbf{82.5} & 1581 \\
        \bottomrule
    \end{tabular}
    \label{table:comp_overall_math}
}
\end{table*}


\section{Experiment Settings}

\paragraph{Training Configuration.} We conduct experiments on GRPO using the veRL framework. To build strong baselines, we adopt several techniques from DAPO, including Clip-Higher and Group-Sampling. Detailed settings are in Appendix \ref{train_hyperparameters}.

\noindent\textbf{Datasets.} Our training data are sourced from DAPO and MATH. To ensure training efficiency, we filter samples that are either too trivial or intractable (see Appendix~\ref{sec:data_filtering} for details). For evaluation, four benchmarks, including MATH-500 \cite{MATH500}, AIME2025, AIME24, Olympiadbench \cite{he2024olympiadbench}, are selected.

\noindent\textbf{Baselines.} We conduct experiments on Qwen2.5-3\/7B and Qwen3-14B. Baselines span three categories: (1)entropy regularization (Entropy-Reg~\cite{chao2024maxentropy} and Entropy-Adv~\cite{entropy_adv}) (2) token-selective methods (KL-Cov~\cite{covclip} and 80/20 \cite{2_8_entropy}), and (3) self-reward methods (SPINE~\cite{wu2025spine} and SRLM~\cite{yuan2024self}). 

\begin{table}[!tbp]
    \centering
    \caption{\textbf{Diversity Analysis on GSM8K Benchmark.} \textbf{Accuracy} is measured by Pass@1. \textbf{Diversity} is evaluated via Distinct-1, Distinct-4, $1-$Self-BLEU, and $1-$Self-ROUGE (higher is better for all metrics). }
    \label{tab:diversity_analysis}
    \vspace{-5pt}
    \setlength\tabcolsep{0.22em}
    
    \resizebox{\linewidth}{!}{
        \renewcommand{\arraystretch}{0.8} 
        \begin{tabular}{l|cccccc}
        \toprule
        \bf{Method} & \textbf{Pass@1} & \textbf{Distinct-1} & \textbf{Distinct-4} & \textbf{Self-BLEU} & \textbf{Self-ROUGE} \\ \hline

        \multicolumn{6}{l}{\gray{\textit{Qwen2.5-3B}}} \\
        Base        & 85.7 & 0.20 & 0.44 & 0.25 & 0.28 \\
        80/20       & 89.1 & 0.26 & 0.48 & 0.30 & 0.36 \\
        \rowcolor{gray!10}Entropy.Reg & 86.8 & 0.32 & \textbf{0.78} & 0.82 & 0.78 \\
        \brow{I\textsuperscript{2}B-LPO} & \textbf{92.8} & \textbf{0.35} & 0.76 & \textbf{0.85} & \textbf{0.81} \\ \hline

        \multicolumn{6}{l}{\gray{\textit{Qwen2.5-7B}}} \\
        Base        & 91.6 & 0.28 & 0.51 & 0.34 & 0.39 \\
        80/20       & 93.5 & 0.35 & 0.65 & 0.38 & 0.45 \\
        \rowcolor{gray!10}Entropy.Reg & 93.8 & 0.57 & 0.73 & 0.47 & \textbf{0.81} \\
        \brow{I\textsuperscript{2}B-LPO} & \textbf{95.6} & \textbf{0.69} & \textbf{0.87} & \textbf{0.56} & 0.76 \\ \hline

        \multicolumn{6}{l}{\gray{\textit{Qwen3-14B}}} \\
        Base        & 92.5 & 0.33 & 0.59 & 0.38 & 0.43 \\
        80/20       & 94.2 & 0.37 & 0.62 & 0.41 & 0.45 \\
        \rowcolor{gray!10}Entropy.Reg & 93.4 & 0.56 & 0.71 & 0.51 & 0.57 \\
        \brow{I\textsuperscript{2}B-LPO} & \textbf{97.8} & \textbf{0.59} & \textbf{0.73} & \textbf{0.57} & \textbf{0.63} \\ 
        
        \bottomrule
        \end{tabular}
    }
\end{table}

\begin{figure*}[!t]
    \centering
    \includegraphics[width=0.93\textwidth]{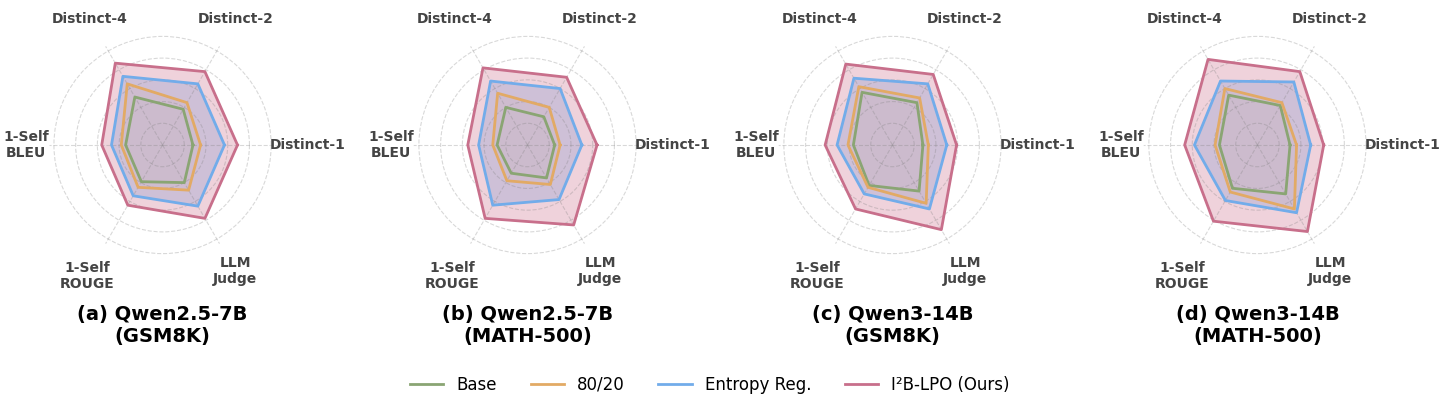}
    \vspace{-5pt}
    \caption{\textbf{The performance of the trained model on six diversity metrics.} We evaluate I\textsuperscript{2}B-LPO using Qwen2.5-7B and Qwen3-14B models across the GSM8K and MATH-500 datasets. For each metric, a higher value indicates greater diversity. And the diversity metrics are calculated across $10$ generated responses per prompt.}
    \label{fig:radar_chart}
\end{figure*}

\begin{figure*}[t] 
    \centering
    \begin{subfigure}[b]{0.68\textwidth}
        \centering
        \includegraphics[width=\linewidth]{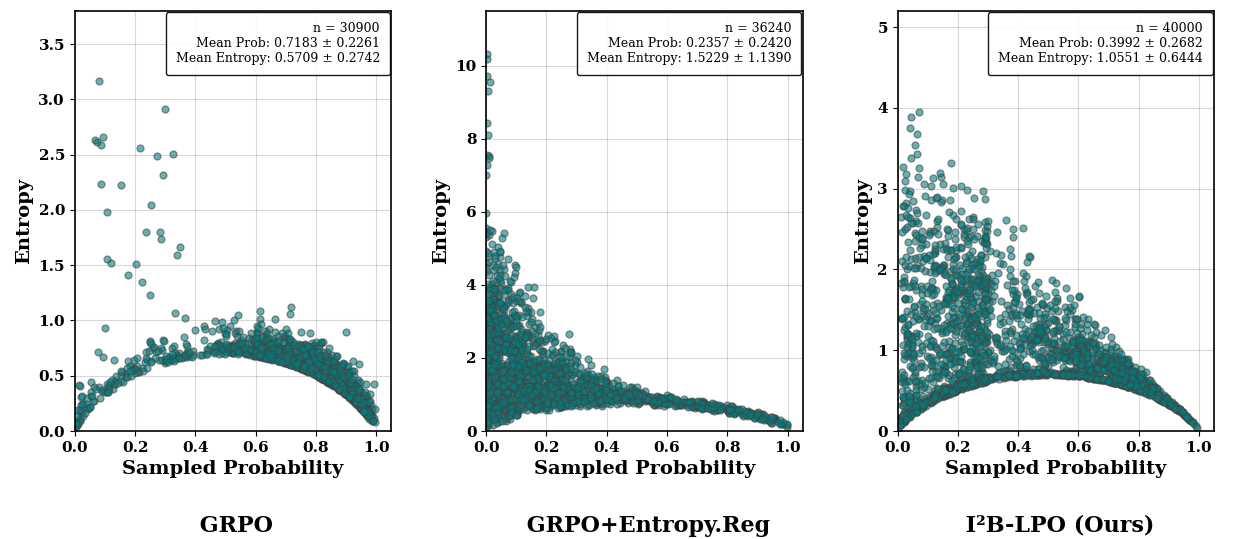}
        \caption{Probability-Entropy scatter plots} 
        \label{fig:entropy_sandian}
    \end{subfigure}
    \hfill 
    \begin{subfigure}[b]{0.30\textwidth}
        \centering
        \scalebox{1}[1.20]{\includegraphics[width=\linewidth]{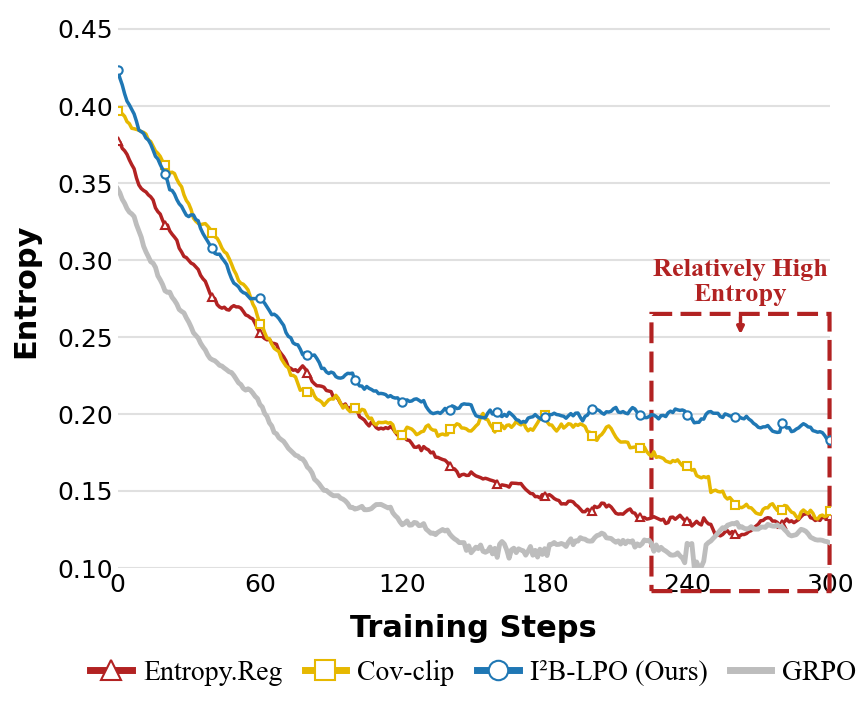}}
        \caption{Entropy Dynamic comparison} 
        \label{fig:entropy}
    \end{subfigure}
    
    \vspace{-0.5em}
    \caption{\textbf{Joint Analysis of Entropy Dynamics.} (a) Probability-Entropy scatter plots of five exploratory tokens from training samples at training step 500 on Qwen2.5-7B-Base, displaying a random sample of 5\% of all data points. (b) Average entropy dynamics across training steps on the MATH dataset.}
    \label{fig:combined_entropy}
\end{figure*}

\noindent\textbf{Metrics.} 
\textbf{\textit{Avg. \# Tokens}} reports the average token length. For diversity, we report Distinct-n~\cite{distinct-n} and Self-BLEU~\cite{bleu}. 
\noindent(1) \textbf{\textit{Distinct-n}} counts the ratio of unique $n$-grams. \noindent(2) \textbf{\textit{Self-BLEU}}. We report $1- \text{Score}$, with higher values indicating greater diversity.\noindent\textbf{\textit{Perplexity (PPL)}} measures uncertainty of the generated sequence(Appendix.\ref{PPL_details}). Low-PPL responses are generally more fluent and semantically coherent.



\section{Experiment}

This section evaluates I\textsuperscript{2}B-LPO's ability to generate diverse, high-quality responses. We aim to answer the following research questions:


\begin{itemize}[leftmargin=*, noitemsep, topsep=0pt, parsep=0pt]
    \item \textbf{RQ1 (Overall Performance):} Does I²B-LPO outperform baseline methods in terms of both reasoning accuracy and semantic diversity?
    \item \textbf{RQ2 (Exploration Behavior Analysis):} How does the entropy-triggered mechanism reshape exploration dynamics? And how does latent injection steer subsequent reasoning---introducing structured exploration or random noise?
    \item \textbf{RQ3 (IB Pruning Efficacy):} How does the IB-based self-reward compare to other self-reward methods, and what specific reasoning characteristics correspond to high and low IB scores?
    \item \textbf{RQ4 (Ablation Study):} How does each component of I²B-LPO contribute to performance? 
\end{itemize}


\subsection{Quality-diversity balance (RQ1)}

In this section, we present fine-grained results on the diversity and quality. 

\noindent\textbf{Quality.} Tab.~\ref{table:comp_overall_math} reports Pass@$n$ ($n \in [1, 256]$) and average response length. I\textsuperscript{2}B-LPO consistently outperforms baselines across both backbones, with advantages widening as $n$ increases. Unlike Entropy-Reg., which artificially inflates entropy via excessive verbosity, I\textsuperscript{2}B-LPO maintains a better balance between quality and efficiency.


\noindent\textbf{Diversity.} We present diversity metrics in Fig.~\ref{fig:radar_chart} and Tab.~\ref{tab:diversity_analysis}. I\textsuperscript{2}B-LPO consistently outperforms baseline models, demonstrating a clear advantage in diversity. While entropy-based methods primarily boost lexical variation (Distinct-$n$), our approach achieves superior performance on semantic metrics (LLM-as-Judge and Self-BLEU).

\subsection{Exploration Behavior Analysis (RQ2)}

This section analyzes the distribution of exploratory tokens and entropy dynamics.


\noindent\textbf{Distributions of Exploratory Tokens.} 
As categorized in Appendix~\ref{sec:token_categories}, high‑entropy tokens fulfill distinct functional roles. Fig.~\ref{fig:combined_entropy} (a) compares the probability–entropy distributions of exploratory tokens (``but”, ``however'', ``thus'', ``wait'', ``let''). The key observations are summarized as follows:
\begin{itemize}[leftmargin=*, noitemsep, topsep=0pt, parsep=0pt]
\item In GRPO, tokens concentrate in high-probability, low-entropy regions, similar to \cite{PathDivergence}.
\item Adding an entropy loss does shift this distribution, but often leads to anomalously high entropy (>8) for some tokens. 
\item I²B-LPO maintains a balanced distribution by activating exploratory tokens across a wider entropy spectrum, preserving their reasoning utility without overfitting into deterministic patterns.
\end{itemize}


\noindent\textbf{Mitigation of Entropy Collapse.} 
Fig.~\ref{fig:combined_entropy} (b) tracks entropy dynamics. While all methods show an initial drop, GRPO suffers from continuous decay, indicating severe entropy collapse. In contrast, I²B-LPO stabilizes entropy levels after the initial phase. This confirms our method effectively preserves exploration capacity, preventing convergence toward deterministic patterns.


\begin{figure*}[!t] 
    \centering
    \includegraphics[width=0.945\textwidth]{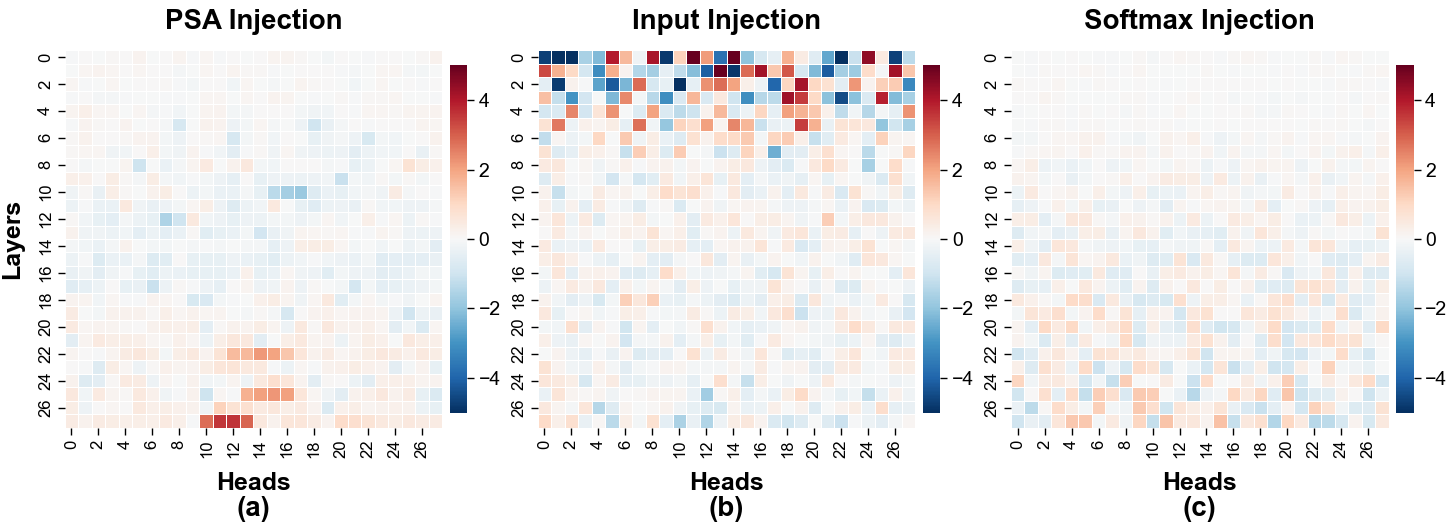}
    \vspace{-8pt}
    \caption{Attention head patterns contrasted between high (Level 9) and low (Level 3) difficulty on Deepmath. Red indicates heads activated by complex problems, while blue denotes heads responsive to simpler ones.}
    \label{fig:heatmaps}
\end{figure*}

\definecolor{MyRed}{HTML}{FA8876}   
\definecolor{MyBlue}{HTML}{6FA8DC} 

\begin{figure*}[!t] 
    \centering
    
    \begin{subfigure}[b]{0.4\linewidth}
        \centering
        \includegraphics[width=\linewidth,height=3.7cm]{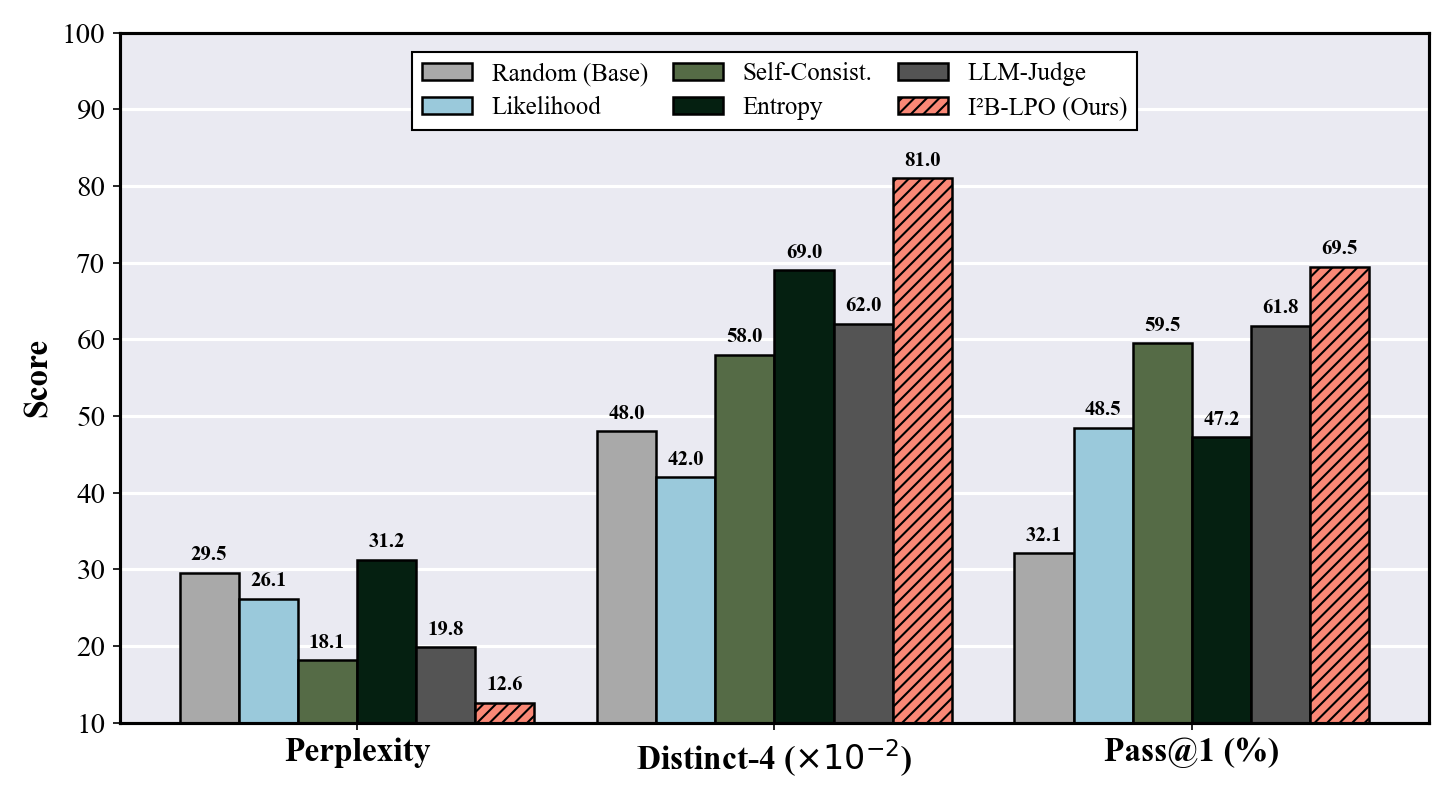}
        \caption{Performance Comparison} 
        \label{fig:bar_chart}
    \end{subfigure}%
    \hfill 
    \begin{subfigure}[b]{0.28\linewidth}
        \centering
        \linespread{1.1}\selectfont 
        \scriptsize 
        \raggedright 
        
        \textbf{Q:} Solve linear equation $2x+4=10$.
        \vspace{0pt}
        
        \begin{tcolorbox}[
        colback=MyRed!10!white,      
        colframe=MyRed,              
        title=\textbf{Low IB (Redundant)}, 
        fonttitle=\bfseries\scriptsize, 
        boxrule=0.5pt,            
        boxsep=2pt, left=2pt, right=2pt, top=2pt, bottom=2pt,
        arc=2pt,
        width=\linewidth
         ]
            \scriptsize 
               \textbf{A:} To solve $2x+4=10$, we subtract 4. $2x=6$. Then divide by 2. So $x=3$. \textit{Let me double check. $2 \times 3 + 4 = 10$. The answer is 3. I think it is 3. Yes, 3.}
     
        \end{tcolorbox}
        
        \vspace{-1.5pt}
        \vspace{-1.0em}
        \begin{tcolorbox}[
        colback=MyBlue!10!white,     
        colframe=MyBlue,             
        title=\textbf{High IB (Concise)}, 
        fonttitle=\bfseries\scriptsize,
        boxrule=0.5pt,
        boxsep=2pt, left=2pt, right=2pt, top=2pt, bottom=2pt,
        arc=2pt,
        width=\linewidth
        ]
            \scriptsize
            \textbf{A:} Subtract 4 from both sides: $2x=6$. Divide by 2: $x=3$. The solution is \boxed{3}.
        \end{tcolorbox}
        \vspace{1pt} 
        \caption{Case Study}
        \label{fig:case_study}
    \end{subfigure}%
    \hfill 
    \begin{subfigure}[b]{0.28\linewidth}
        \centering
        \includegraphics[width=\linewidth,height=3.7cm]{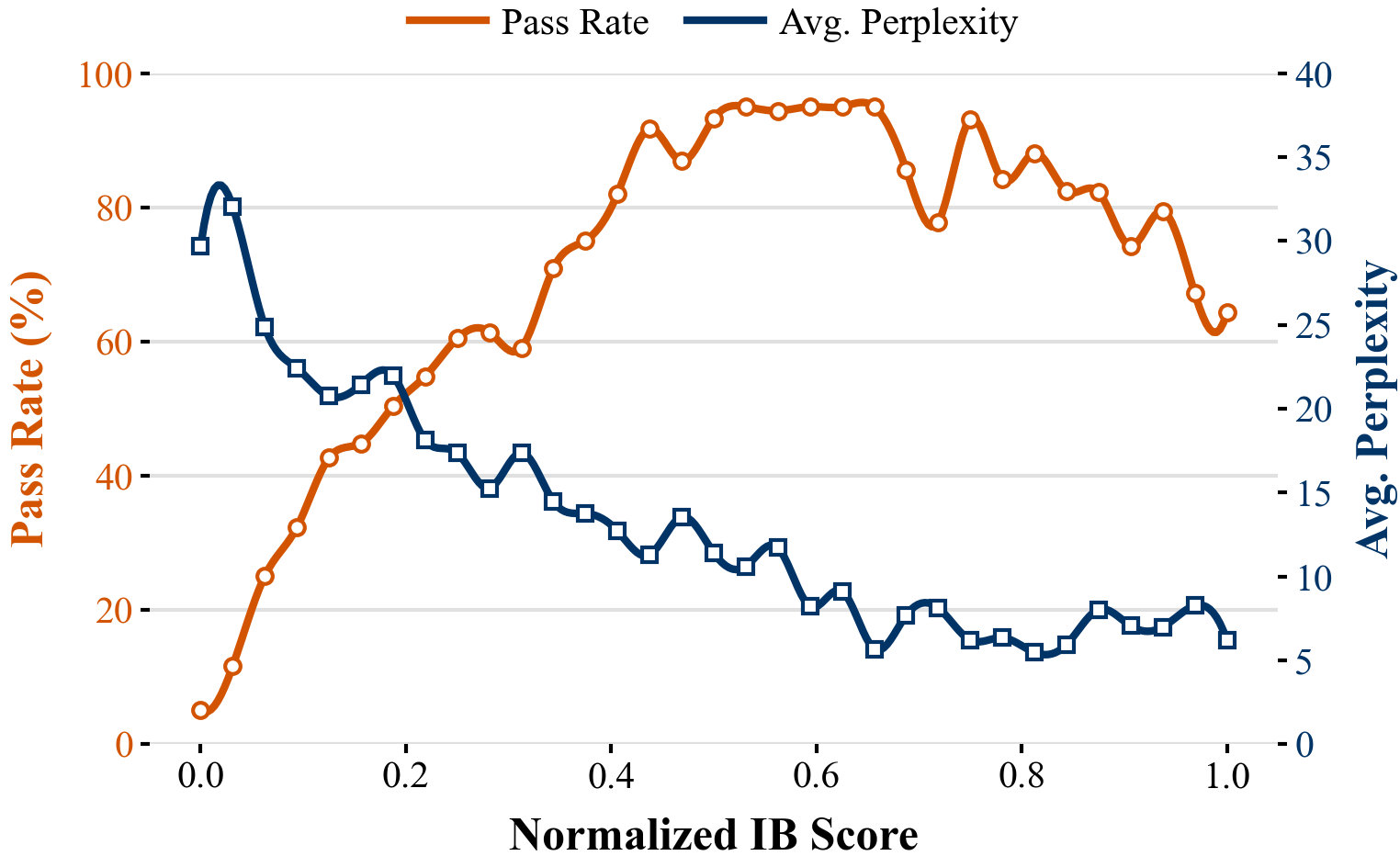}
        \caption{Accuracy \& Perplexity Trends}
        \label{fig:IB_dual_curve}
    \end{subfigure}

    \vspace{-0.2em}
    \caption{\textbf{Overview of Information Bottleneck (IB) Analysis.} (a) Performance comparison on OlympiadBench. (b) Case study illustrating how higher IB scores correlate with more concise reasoning. (c) The impact of IB scores on model accuracy and perplexity.}
    \label{fig:IB_combined_analysis}
    \vspace{-0.3em}
\end{figure*}

\noindent\textbf{Ablation Study of Latent Injection}\label{Latent_Injection} To verify whether latent injection fosters structured exploration rather than introducing random noise. We investigate three injection strategies: 
Early Fusion at input-level, Deep Fusion via PSA (detailed in Sec .~\ref {PSA_guidance}), and Late Fusion at Softmax-layer. The formulations are described below:

\noindent\textcircled{1} \textit{Early Fusion at input-level:} We inject the latent code $z_i$ by adding it element-wise to the input embedding $h(x_t)$ of each token $x_t$. This broadcasts the prior across the sequence dimension:

\begin{equation}
\small
h_i'(x_t) = h(x_t) + z_i, \quad \forall t = 1, \dots, T.
\end{equation}

\noindent\textcircled{2} \textit{Late Fusion at Softmax-layer:} This strategy directly utilizes the projection of $z_i$ to influence the distribution of LLM's vocabulary. We first map the latent vector $z_i$ to a logit adjustment vector $p_{z_i} \in \mathbb{R}^V$. This $p_{z_i}$ is then superimposed onto the original logits $p$ to derive the final distribution:

\begin{equation}
\small
p_i' = p + p_{z_i}, \quad \hat{y} = \text{softmax}(p_i'),
\end{equation}
where $p_{z_i}$ biases the token selection probability $\hat{y}$.

\noindent\textit{Analysis.} Tab.~\ref{tab:ablation_full}.A presents ablation results of
latent injection on qwen2.5-7B. Fig.~\ref{fig:heatmaps} contrasts attention patterns for high- (Level 9) and low-difficulty (Level 3) problems on DeepMath (details in Appendix~\ref{heatmap_details}). Key observations are:

\begin{itemize}[leftmargin=*, noitemsep, topsep=0pt, parsep=0pt]
\item  Input Injection: chaotic early activations diminish in deeper layers, diluted before reasoning.

\item Softmax Injection: modifying logits externally hinders gradients from shaping internal attention, leading to scattered activations.

\item PSA induces ``Structured Activation'': it mobilizes difficulty‑sensitive heads in final layers (24--27), consistent with \cite{heatmap_conclusion,SemanticEntropy}.
\end{itemize}
This confirms that PSA effectively engages deep reasoning without aimless randomness.




\subsection{Self-Reward Effectiveness (RQ3)}

In this section, we benchmark IB metric against other self-rewards methods: Likelihood~\cite{Likelihood}, Self-Consistency~\cite{wang2022self}, Entropy.Reg~\cite{chao2024maxentropy}, and LLM-Judge~\cite{zheng2024judging}. As shown in Fig.~\ref{fig:IB_combined_analysis}(\subref{fig:bar_chart}), IB self-reward outperforms baselines across accuracy, diversity, and confidence (perplexity). Notably, it resolves the diversity-confidence trade-off observed in Entropy.Reg, which sacrifices confidence for diversity. 
Fig.~\ref{fig:IB_combined_analysis}(\subref{fig:case_study})
shows that higher IB scores correspond to concise reasoning chains, while lower scores identify redundant content. 
Fig.~\ref{fig:IB_combined_analysis}(\subref{fig:IB_dual_curve})
further demonstrates that IB metric balances exploration and confidence by reducing perplexity while improving quality.

\begin{table*}[t]
\centering
\scriptsize
\setlength{\tabcolsep}{5.5pt} 
\caption{\textbf{Ablation Study of I²B-LPO Components on Qwen2.5-7B.} We evaluate the contribution of core components: Branching Triggers, Latent Injection, and IB self-rewards. To isolate the impact of topological structures, Blocks A and B are conducted without the IB self-reward mechanism.  \textbf{Note:} non-ablated components are fixed to optimal settings( $K=7$, $N=8$). $^\dagger$ denotes the PSA Fusion without the injection weight decay.}
\label{tab:ablation_full}
\renewcommand{\arraystretch}{0.90} 
\begin{tabular}{l ccc ccc ccc ccc}
\toprule
\multirow{3}{*}{\textbf{Method}} 
& \multicolumn{3}{c}{\textbf{AIME25}} 
& \multicolumn{3}{c}{\textbf{AIME24}} 
& \multicolumn{3}{c}{\textbf{MATH}} 
& \multicolumn{3}{c}{\textbf{OlympiadBench}} \\
\cmidrule(lr){2-4} \cmidrule(lr){5-7} \cmidrule(lr){8-10} \cmidrule(lr){11-13}
& \textbf{Pass@1} & \textbf{Dist-4} & \textbf{PPL} 
& \textbf{Pass@1} & \textbf{Dist-4} & \textbf{PPL}
& \textbf{Pass@1} & \textbf{Dist-4} & \textbf{PPL}
& \textbf{Pass@1} & \textbf{Dist-4} & \textbf{PPL} \\

\midrule
\rowcolor{gray!10} \textbf{GRPO Baseline} & 8.2 & 0.16 & 32.2 & 10.3 & 0.19 & 27.8 & 54.4 & 0.40 & 17.6 & 44.9 & 0.32 & 19.8 \\
\midrule
\arrayrulecolor{Exp}
\addlinespace[-2.6pt]
\specialrule{1.7pt}{0pt}{0pt} 
\rowcolor{Exp}
\multicolumn{13}{c}{\textbf{A: Latent Injection Ablation} (Fixed: Entropy Branching+ w/o IB)}\\
\specialrule{1.5pt}{0pt}{0pt}
\arrayrulecolor{black}
\addlinespace[-1.9pt]
\midrule
w/o Latent Injection & 10.2 & 0.34 & 32.2 & 15.5 & 0.39 & 27.8 & 59.7 & 0.64 & 18.4 & 48.3 & 0.50 & 19.5 \\
+ Input-Level Fusion & 10.6 & 0.38 & 30.5 & 16.2 & 0.48 & 26.5 & 61.2 & 0.68 & 16.2 & 49.5 & 0.52 & 19.8 \\
+ Softmax-Level Fusion & 10.8 & 0.41 & 38.6 & 16.4 & 0.45 & 33.4 & 63.5 & 0.66 & 21.5 & 50.8 & 0.55 & 25.4 \\
\rowcolor{OursRow}  + Norm Modulation &  11.0 & 0.43 & 36.5 &17.0 & 0.46 & 31.8 &64.2 & 0.72 & 20.8 & 51.2 & 0.56 & 24.5 \\
\rowcolor{OursRow}  + KV-Augmentation & 11.2 & 0.44 & 35.8&17.2 & 0.47 & 31.0 & 65.1 & 0.74& 20.1 &51.6 &0.57 & 23.8 \\
\rowcolor{OursRow} + PSA Fusion$^\dagger$ & 11.6 & 0.45 & 35.1 & 17.2 & 0.48 & 30.5 & 65.4 & 0.75 & 19.2 & 51.9 & 0.57 & 23.0 \\
\rowcolor{OursRow} \textbf{+ PSA Fusion (Ours)} & 11.8 & 0.47 & 34.2 & 17.5 & 0.49 & 29.8 & 66.7 & 0.78 & 19.5 & 52.3 & 0.59 & 22.1 \\
\textit{$\Delta$ vs. w/o Latent Injection} 
                  & \pair{\inc{1.6}}{\inc{0.13}}              
                  & \pair{\textcolor{red}{+2.0}}{\inc{2.0}}   
                  & \pair{\inc{0.1}}{\textcolor{red}{+2.0}}           
                  & \pair{\inc{0.1}}{\inc{0.14}}              
                  & \pair{\textcolor{red}{+1.1}}{\inc{4.0}}   
                  & \pair{\inc{0.09}}{\textcolor{red}{+2.6}}  
                  \\
\midrule
\arrayrulecolor{Exp}
\addlinespace[-2.6pt]
\specialrule{1.7pt}{0pt}{0pt} 
\rowcolor{Exp}
\multicolumn{13}{c}{\textbf{B: Branching Trigger Ablation} (Fixed: PSA + w/o IB)}\\
\specialrule{1.5pt}{0pt}{0pt}
\arrayrulecolor{black}
\addlinespace[-1.9pt]
\midrule
w/o Branching ($K=0$) & 8.6 & 0.21 & 30.1 & 13.4 & 0.23 & 26.7 & 54.5 & 0.43 & 17.3 & 45.3 & 0.35 & 19.1 \\
+ Random Branching & 9.1 & 0.24 & 31.9 & 13.6 & 0.34 & 25.9 & 55.4 & 0.46 & 17.8 & 46.7 & 0.40 & 20.5 \\
+ Likelihood Branching & 9.4 & 0.26 & 32.0 & 14.2 & 0.37 & 27.3 & 56.7 & 0.51 & 18.1 & 47.9 & 0.45 & 19.8 \\
\rowcolor{OursRow}  \textbf{+ Entropy Branching} 
& 10.2 & 0.34 & 32.2 & 15.7 & 0.39 & 27.8 & 59.7 & 0.64 & 18.4 & 48.3 & 0.50 & 19.5 \\

\textit{$\Delta$ vs. w/o Branching} 
                  & \pair{\inc{1.6}}{\inc{0.13}}              
                  & \pair{\textcolor{red}{+2.1}}{\inc{2.3}}   
                  & \pair{\inc{0.16}}{\textcolor{red}{+1.1}}  
                  & \pair{\inc{5.2}}{\inc{0.21}}              
                  & \pair{\textcolor{red}{+1.1}}{\inc{3.0}}   
                  & \pair{\inc{0.15}}{\textcolor{red}{+0.4}}  
                  \\
\midrule
\arrayrulecolor{Exp}
\addlinespace[-2.6pt]
\specialrule{1.7pt}{0pt}{0pt} 
\rowcolor{Exp}
\multicolumn{13}{c}{\textbf{C: IB Mechanism Ablation} (Fixed: Entropy + PSA)}\\
\specialrule{1.5pt}{0pt}{0pt}
\arrayrulecolor{black}
\addlinespace[-1.9pt]
\midrule
w/o IB  & 11.8 & 0.47 & 34.2 & 17.5 & 0.49 & 29.8 & 66.7 & 0.78 & 19.5 & 52.3 & 0.59 & 22.1 \\
+ IB Loss & 12.2 & 0.49 & 24.1 & 18.1 & 0.52 & 20.4 & 78.3 & 0.76 & 14.2 & 54.6 & 0.60 & 14.5 \\
+ IB Pruning  & 12.9 & 0.50 & 25.0 & 18.3 & 0.51 & 25.4 & 79.6 & 0.80 & 15.6 & 56.8 & 0.65 & 16.5 \\
\midrule
\rowcolor{OursRow} \textbf{I²B-LPO (Full Method)} 
 & \textbf{13.6} & \textbf{0.51} & 22.5 & \textbf{18.6} & \textbf{0.54} & 19.2 & \textbf{81.5} & \textbf{0.82} & 13.1 & \textbf{58.0} & \textbf{0.65} & 12.6 \\
\textit{$\Delta$ vs. w/o IB} 
                  & \pair{\inc{1.8}}{\inc{0.04}}              
                  & \pair{\textcolor{ForestGreen}{\textbf{-11.7}}}{\inc{1.1}}  
                  & \pair{\inc{0.05}}{\textcolor{ForestGreen}{\textbf{-10.6}}} 
                  & \pair{\inc{14.8}}{\inc{0.04}}             
                  & \pair{\textcolor{ForestGreen}{\textbf{-6.4}}}{\inc{5.7}}   
                  & \pair{\inc{0.06}}{\textcolor{ForestGreen}{\textbf{-9.5}}}  
                  \\
\bottomrule
\end{tabular}
\vspace{-10pt}
\end{table*}

\definecolor{HparamBlue}{HTML}{E6F2FF}   
\definecolor{HparamGreen}{HTML}{E8F5E9}  
\definecolor{HparamYellow}{HTML}{FFFDE7} 
\definecolor{BestRow}{HTML}{E0E0E0}      

\begin{table}[!tbp]
\centering
\footnotesize 
\setlength{\tabcolsep}{6.0pt} 
\caption{\textbf{Hyperparameter Sensitivity Analysis.} 
\colorbox{HparamBlue}{Blue} indicates Injection Depth; 
\colorbox{HparamGreen}{Green} indicates Branching Factor ($K$); 
\colorbox{HparamYellow}{Yellow} indicates Max Response Length. 
The bold rows denote the chosen settings.}
\label{tab:hyperparam_color}
\vspace{-2pt}
\scalebox{0.96}{
\renewcommand{\arraystretch}{0.8} 
\begin{tabular}{l cc cc}
\toprule
\multirow{2}{*}{\textbf{Configuration}} 
& \multicolumn{2}{c}{\textbf{MATH}} 
& \multicolumn{2}{c}{\textbf{OlympiadBench}} \\
\cmidrule(lr){2-3} \cmidrule(lr){4-5}
& \textbf{Pass@1} & \textbf{Dist-4} 
& \textbf{Pass@1} & \textbf{Dist-4} \\
\midrule

\multicolumn{5}{l}{\textcolor{gray}{\textit{PSA Injection Depth}}} \\
\rowcolor{HparamBlue} Last 4 Layers  & 76.5 & 0.75 & 53.2 & 0.58 \\
\rowcolor{HparamBlue} Last 8 Layers  & 79.8 & 0.79 & 56.4 & 0.62 \\
\rowcolor{HparamBlue} \textbf{Last 12 Layers} & \textbf{81.5} & \textbf{0.82} & \textbf{58.0} & \textbf{0.65} \\
\rowcolor{HparamBlue} Last 16 Layers & 81.1 & 0.81 & 57.6 & 0.64 \\
\rowcolor{HparamBlue} Last 20 Layers & 80.4 & 0.80 & 56.9 & 0.63 \\
\rowcolor{HparamBlue} Last 28 Layers & 78.2 & 0.78 & 55.1 & 0.61 \\

\multicolumn{5}{l}{\textcolor{gray}{\textit{Branching Factor}}} \\
\rowcolor{HparamGreen} K=1 & 75.5 & 0.75 & 52.8 & 0.54 \\
\rowcolor{HparamGreen} K=3 & 78.7 & 0.78 & 55.8 & 0.60 \\ 
\rowcolor{HparamGreen} \textbf{K=7} & \textbf{81.5} & \textbf{0.82} & \textbf{58.0} & \textbf{0.65} \\
\rowcolor{HparamGreen} K=15 & 82.8 & 0.86 & 58.4 & 0.69 \\

\multicolumn{5}{l}{\textcolor{gray}{\textit{Max Response Length}}} \\
\rowcolor{HparamYellow} 2048 Tokens & 68.2 & 0.70 & 45.5 & 0.55 \\
\rowcolor{HparamYellow} 4096 Tokens & 79.5 & 0.80 & 56.2 & 0.63 \\
\rowcolor{HparamYellow} \textbf{8192 Tokens} & \textbf{81.5} & \textbf{0.82} & \textbf{58.0} & \textbf{0.65} \\

\bottomrule
\end{tabular}
} 
\vspace{-10pt}
\end{table}

\subsection{Ablation of Main Components (RQ4)}\label{Ablation_main}


Tab.~\ref{tab:ablation_full} presents ablation results on the core components. The results indicate that (1) entropy-based branching outperforms other strategies by accurately targeting high-uncertainty nodes; (2) the full IB mechanism (loss + pruning) forms an explore–converge loop, significantly reducing PPL by 11.7. (3) Overall, the complete framework elevates MATH Pass@1 from 54.4\% to 81.5\%, maintaining high diversity and low perplexity. Tab. \ref{tab:hyperparam_color} shows ablation results of key hyperparameters. PSA injection depth follows an inverted U-curve, peaking at the last 12 layers. The max\_length of 8192 tokens provides sufficient capacity for complex reasoning.



 


\section{Conclusion}
This paper addresses the exploration collapse caused by the semantic homogeneity of reasoning paths. Our I²B-LPO framework enables structured exploration via entropy-driven latent branching and a dual-purpose IB for simultaneous filtering and reward shaping. Extensive experiments confirm that I²B-LPO achieves SOTA accuracy and diversity, effectively overcoming the exploration bottleneck. This work provides a principled and extensible framework for exploration-aware policy optimization in LLM reasoning.

\section*{Limitations and Potential Risk}
 \noindent\textbf{Limitations.} First, entropy as an uncertainty measure may not consistently reflect semantic diversity in open-ended or non-mathematical tasks. Second, despite IB-based pruning, the branching mechanism remains more computationally intensive than single-rollout approaches. We aim to explore more efficient algorithms in the future.
 
 \noindent\textbf{ Potential Risk.} Our method focuses on enhancing mathematical reasoning capabilities using established public benchmarks. As the proposed approach operates strictly within abstract symbolic domains and does not involve sensitive data or content generation, it poses minimal ethical risks or safety concerns.

\section*{Ethical Considerations}
This study adheres to the ACL Code of Ethics and has considered relevant ethical issues throughout the research process. All data are sourced from publicly available datasets and have been anonymized to protect privacy. The experimental process and results are reported truthfully, and reproducible code and resources are provided. We recommend implementing oversight and filtering mechanisms in practical applications. The authors have completed and submitted the Responsible NLP Research Checklist and are committed to maintaining integrity and transparency in the research.





\FloatBarrier

\bibliography{main}

\appendix

\clearpage
\section*{Appendix}

\section{Categorizing High-Entropy Tokens
}\label{sec:token_categories}

\begin{figure}[htbp]
    \centering
    \includegraphics[width=0.95\linewidth]{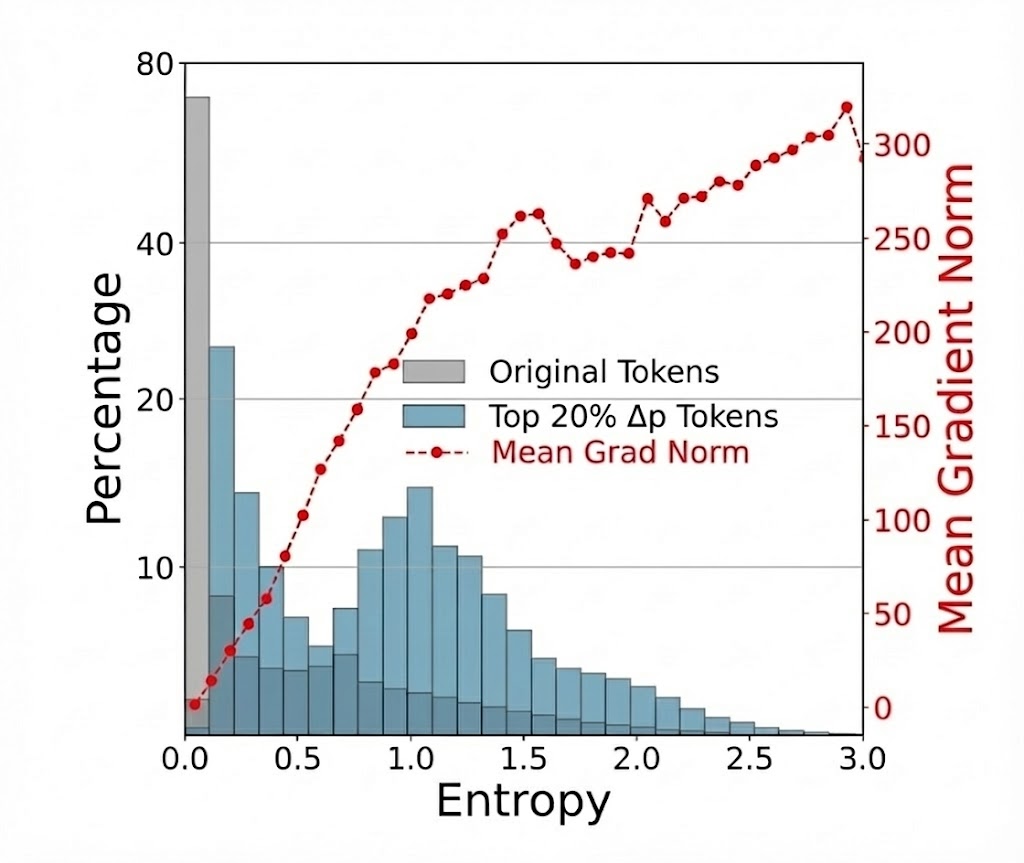}
    \caption{Token entropy and gradient
distribution.}
    \label{fig:entropy_gradient}
\end{figure}

In Fig. \ref{fig:entropy_gradient}, Top20\% impactful updates primarily target high-entropy
tokens. These tokens tend to produce larger gradients during backpropagation. This indicates that progress in this stage is mainly driven by resolving uncertainty at critical “forks” in reasoning paths \cite{2_8_entropy}. In RLVR, tokens generated by models exhibit different functional roles that collectively drive the reasoning process. Based on their operational characteristics, we categorize tokens into three roles:

\begin{itemize}[leftmargin=*, noitemsep, topsep=0pt, parsep=0pt]
    \item \textbf{Logical Structuring Tokens:} Govern reasoning flow (\eg\ causal, contrastive, progressive, and parallel connectors). They help structure multi-step argumentation or explanations.
    \item \textbf{Metacognitive Tokens:} Reflect meta-cognitive functions, especially self-monitoring behaviors (\eg\ verifying, summarizing, and revising). These tokens actively guide the reasoning process through reflective adjustment and solution refinement.
    \item \textbf{Semantic Support Tokens:} Provide linguistic elements that ensure fluency, coherence, and informativeness (\eg\ core grammatical elements, domain-specific entities).
\end{itemize}

\noindent We provide examples of each category in Table~\ref{tab:token_examples}.

\begin{table}[ht]
\centering
\begin{tabular}{l p{0.50\linewidth}}
\toprule
\textbf{Category} & \textbf{Examples} \\
\midrule
\addlinespace[1ex] 
\textbf{Logical Structuring} & Causal (\eg\ `therefore', `because'), contrastive (\eg\ `however', `but'), progressive (\eg\ `first', `next', `finally'), and parallel (\eg\ `and', `also'). \\
\addlinespace[1ex]
\textbf{Metacognitive} & Verifying (\eg\ `Let's check'), revising (\eg\ `Correction', `Wait'), summarizing (\eg\ `In summary'), and planning (\eg\ `First, I will...'). \\
\addlinespace[1ex]
\textbf{Semantic Support} & Grammatical elements (\eg\ `the', `is', `of'), domain entities (\eg\ `problem', `solution'), and adjectives (\eg\ `correct', `final'). \\
\bottomrule
\end{tabular}
\caption{Examples of Token Categories in RLVR.}
\label{tab:token_examples}
\end{table}

\section{Implementation Details of the CVAE}\label{CVAE_details}

In this section, we provide the architectural specifications and the training objective of our CVAE.
\noindent\textbf{Network Architecture:} A typical CVAE consists of three parts: (1) The \textbf{Encoder} $q_\phi(z\mid x, c)$ approximates the posterior distribution of the latent variable $z$. (2) The \textbf{Prior Network} $p_\theta(z\mid c)$ models the prior distribution of $z$, which is conditioned only on the context $c$. (3) The \textbf{Decoder Network} $p_\theta(y\mid z, x)$ takes $z$ and $x$ to decode the output $y$.

Specifically, we employ \textbf{DeBERTa-v2-large}~\citep{he2020deberta} as the Encoder $q_\phi(z|x,c)$. DeBERTa is chosen for its disentangled attention mechanism, which effectively encodes both content and relative positions. Formally, given the input sequence $x$, we first extract the contextualized hidden states via the encoder. To obtain a fixed-size vector representation $h_x \in \mathbb{R}^{d_{model}}$, we apply mean pooling over the token embeddings:
\begin{equation}
    h_x = \text{MeanPool}(\text{DeBERTa}(x)),
\end{equation}
where $d_{model} = 1024$ for the large architecture. To construct the probabilistic latent space, $h_x$ is projected into the variational parameters---mean $\mu$ and log-variance $\log \sigma^2$---via two separate linear transformations:
\begin{equation}
    \mu = W_\mu h_x + b_\mu, \quad \log \sigma^2 = W_\sigma h_x + b_\sigma,
\end{equation}
where $W_{(\cdot)} \in \mathbb{R}^{d_z \times d_{model}}$ are learnable projection matrices and $d_z$ is the dimension of the latent bottleneck (e.g., $d_z=128$). Finally, the latent variable $z$ is sampled using the standard reparameterization trick:
\begin{equation}
    z = \mu + \sigma \odot \epsilon, \quad \epsilon \sim \mathcal{N}(0, I).
\end{equation}
This latent code $z$ serves as a compact semantic anchor, which is subsequently fused into the decoder (LLaMA) to guide the generation of output $y$.

\noindent\textbf{Training Objective} Formally, the CVAE is trained by maximizing the Evidence Lower Bound:
\begin{equation}
\small
\begin{aligned}
    L_{\text{ELBO}} &= L_{\text{REC}} - L_{\text{KL}} \\
    &= \mathbb{E}_{q_\phi(z | x, y)} [\log p_\theta(y | z, x)] \\
    &\quad - \text{KL}(q_\phi(z | x, y) \| p_\theta(z | x)) \\
    &\leq \log p(y | x),
\end{aligned}
\end{equation}
where $ L_{\text{REC}} $ denotes the reconstruction loss, and $ L_{\text{KL}} $ represents the Kullback-Leibler divergence between the posterior and prior distributions. The CVAE is trained on MATH and GSM8K training sets.

\section{IB-Aware Scoring Metric}\label{IB_details}

\paragraph{General Theory.}\label{IB_Theory}

Information Bottleneck (IB) formalizes the trade-off between compressing input $X$ into a compact representation $\hat{X}$ while maintaining predictive power for target $Y$, via the Lagrangian objective:
\begin{equation}
\label{eq:ib_loss}
\mathcal{L}_{\text{IB\_naive}} = I(X;\hat{X}) - \beta I(\hat{X};Y),
\end{equation}
where $I(\cdot;\cdot)$ denotes Mutual Information (MI) and $\beta$ controls the trade-off.
The MI between $X$ and $Y$ is defined as:
\begin{equation}
\small
I(X;Y) = H(X) - H(X|Y),
\label{MI}
\end{equation}
where $H(\cdot)$ denotes entropy. A high value of $I(X;Y)$ indicates that knowledge of one variable reduces uncertainty about the other. 

\paragraph{Derivation of the IB-Score.}\label{IB_Derivation}

Following the \textit{IB-Aware Reasoning Optimization}, the optimization objective in LLM reasoning is formulated as:
\begin{equation}
     \mathcal{L}_{\text{IB}} = { {I(q; r) -\beta  I(r; a)}}.
\end{equation}
To derive a tractable metric, we decompose these two terms respectively.

\noindent\textbf{Minimizing Complexity $I(q; r)$.}
We first expand $I(q; r)$ using the definition of mutual information and the assumption that the marginal entropy of reasoning $H(r)$ is invariant under the policy $\pi$ (Assumption 1 in \cite{lei2025revisiting}). Applying the chain rule of entropy to the autoregressive trajectory $r = (o_1, \dots, o_T)$, we obtain:
\begin{equation}
\begin{aligned}
    I(q; r) &= H(r) - H(r \mid q) \\
            &= H(r) - \sum_{t=1}^{T} H(o_t \mid o_{<t}, q).
\end{aligned}
\end{equation}
Therefore, minimizing the mutual information $I(q; r)$ is equivalent to maximizing the conditional entropy sum $\sum_{t} H(o_t \mid o_{<t}, q)$. This term encourages the model to maintain high entropy during generation, preventing it from collapsing into memorized or over-deterministic patterns dependent solely on the specific prompt phrasing.

\noindent\textbf{Maximizing Informativeness $I(r; a)$.}
For the second term, we have $I(r; a) = H(r) - H(r \mid a)$. Maximizing this quantity requires minimizing the conditional entropy $H(r \mid a)$. 
In our specific selection phase, we restrict our candidate pool to the validity-verified subset $\mathcal{R}_{correct} = \{r \in \mathcal{R} \mid \hat{a}(r) = a\}$. For any path $r \in \mathcal{R}_{correct}$, the reasoning $r$ implies the answer $a$ with certainty (i.e., $p(a|r) \approx 1$). 
Under this condition, the posterior probability of the reasoning path answered can be approximated by the likelihood of the path itself:
\begin{equation}
    p(r \mid q, a) = \frac{p(a \mid r, q) p(r \mid q)}{p(a \mid q)} \propto p(r \mid q).
\end{equation}
Consequently, minimizing the uncertainty of the reasoning answered ($H(r \mid q, a)$) corresponds to selecting trajectories that maximize the log-likelihood $\log \pi(r \mid q)$. High-likelihood paths within $\mathcal{R}_{correct}$ represent the most confident reasoning traces that lead to correct solution.

\noindent\textbf{The IB-Score formulation.}
Combining the derivations above, we transform the trajectory-level IB objective into a tractable token-level utility function. By substituting the log-likelihood for the informativeness term and the token-level entropy for the complexity term, we propose the \textbf{IB-Score} for a reasoning path $r$:

{\small
\begin{equation} 
\label{eq:ib_derivation_correct}
\begin{split}
    \mathcal{S}_{\text{IB}}(r) 
    &= \sum_{t=1}^{T} \left( 
        \underbrace{H(o_t \mid o_{<t}, q)}_{\substack{\text{Exploration Proxy} \\ (\text{Max } I(q;r))}} 
        - 
        \underbrace{\beta H(o_t \mid o_{<t}, q, a)}_{\substack{\text{Relevance Proxy} \\ (\text{Min } I(r;a))}}
    \right) \\
    &= \sum_{t=1}^{T} \lambda_t H(o_t \mid o_{<t}, q) \\
    &\cong \sum_{t=1}^{T} \mathcal{A}_t H(o_t \mid o_{<t}, q)
\end{split}
\end{equation}
}

We formulate the Information Bottleneck objective as maximizing a score $\mathcal{S}_{\text{IB}}$. This score balances exploration (Standard Entropy) against relevance (Negative Conditional Entropy):
{\small
\begin{equation} 
\label{eq:ib_score_derivation}
\begin{split}
    \mathcal{S}_{\text{IB}}(r) 
    &= \sum_{t=1}^{T} \underbrace{\left( H(o_t \mid o_{<t}, q) - \beta H(o_t \mid o_{<t}, q, a) \right)}_{s_t^{\text{IB}} \text{: Per-token IB Score}} \\
    &= \sum_{t=1}^{T} s_t^{\text{IB}}
\end{split}
\end{equation}
}

Let $H_t = H(o_t \mid o_{<t}, q)$ and $H_{t|a} = H(o_t \mid o_{<t}, q, a)$.
We set $\beta=2$ as per the theoretical analysis. The per-token score becomes:
\begin{equation}
    s_t^{\text{IB}} = H_t - 2 H_{t|a}
\end{equation}
Using the inequality $0 \le H_{t|a} \le H_t$, we analyze the bounds of $s_t^{\text{IB}}$:
\begin{align}
    &\text{Best:} && H_{t|a}=0 \implies s_t^{\text{IB}} = \mathbf{H_t} \\
    &\text{Worst:} && H_{t|a}=H_t \implies s_t^{\text{IB}} = \mathbf{-H_t}
\end{align}
Thus, the score term lies in the symmetric interval:
\begin{equation}
    s_t^{\text{IB}} \in [-H_t, H_t]
\end{equation}
We can therefore represent $s_t^{\text{IB}}$ as a modulated entropy term:
\begin{equation}
    s_t^{\text{IB}} = \lambda_t H_t, \quad \text{where } \lambda_t \in [-1, 1]
\end{equation}
Now we map the coefficient $\lambda_t$ to the Reinforcement Learning context:
\begin{itemize}
    \item \textbf{High Score ($\lambda_t \to 1$)}: Occurs when the token is perfectly predictive. This corresponds to a "Good" action.
    \item \textbf{Low Score ($\lambda_t \to -1$)}: Occurs when the token provides no information about the answer. This corresponds to a "Bad" action.
\end{itemize}
The \textbf{Advantage} function $\mathcal{A}_t$ naturally captures this property: high positive $\mathcal{A}_t$ for good actions, negative $\mathcal{A}_t$ for bad actions. 
Thus, we can directly approximate $\lambda_t$ with the advantage (without any negative sign):
\begin{equation}
    \lambda_t \approx \mathcal{A}_t
\end{equation}
Substituting $\lambda_t \approx \mathcal{A}_t$ back into the summation:
\begin{equation}
\begin{split}
    \mathcal{S}_{\text{IB}}(r) 
    &= \sum_{t=1}^{T} s_t^{\text{IB}} \\
    &= \sum_{t=1}^{T} \lambda_t H_t \\
    &\cong \sum_{t=1}^{T} \mathcal{A}_t H(o_t \mid o_{<t}, q)
\end{split}
\end{equation}

\section{Localization of Difficulty-Sensitive Attention Heads}\label{heatmap_details}

Fig.~\ref{fig:heatmaps} depicts attention head patterns contrasting high (Level 9) vs. low (Level 3) difficulty on DeepMath~\cite{he2025deepmath}, a dataset spanning 13 difficulty levels (3.0 to 9.0). Let $H \in \mathbb{R}^{B \times L \times N \times d}$ denote the output tensor of the multi-head attention (MHA) layer, where $B$ is the batch size, $L$ is the sequence length, $N$ is the number of attention heads, and $d$ is the head dimension.

The final contextual representation $Z$ is typically obtained by projecting the concatenated output of all heads via the output projection matrix $W_o \in \mathbb{R}^{(Nd) \times D}$:
\begin{equation}
Z = \text{Reshape}(H) W_o^\top \in \mathbb{R}^{B \times L \times D},
\end{equation}
where $D = N \times d$ represents the model's hidden dimension.

To analyze the independent contribution of the $i$-th attention head ($i \in \{1, \dots, N\}$), an ablation strategy is employed. An ablated representation $H^{(i)}$ is constructed by retaining the output of the $i$-th head while zeroing out all other heads:


\begin{equation}
\begin{split}
H_{b,\ell,j,:}^{(i)} = 
\begin{cases} 
H_{b,\ell,i,:} & \text{if } j = i \\
\mathbf{0} & \text{otherwise}
\end{cases}, \\
\forall b \in [B], \ell \in [L], j \in [N]
\end{split}
\end{equation}

Subsequently, the projected representation $Z^{(i)}$ is computed using $W_o$. We extract the embedding of the \textbf{last token} for each sample $b$, denoted as $z_b^{(i)} = Z_{b, L-1, :}^{(i)} \in \mathbb{R}^D$.

\subsection{Difficulty Scoring}
Using the pre-trained probe direction $v_{diff}$, the difficulty score contribution of the $i$-th head for sample $b$ is calculated as the normalized projection onto the difficulty direction:
\begin{equation}
s_b^{(i)} = \frac{\langle z_b^{(i)}, v_{diff} \rangle}{\|v_{diff}\|_2}.
\end{equation}

For a batch of samples sharing the same ground-truth difficulty level, the mean head-wise attribution is aggregated:
\begin{equation}
\bar{s}^{(i)} = \frac{1}{B} \sum_{b=1}^{B} s_b^{(i)}.
\end{equation}

\subsection{Differentiation Score and Localization}
To determine whether a specific head is sensitive to hard or easy problems, the \textit{Differentiation Score} is computed as the difference between the mean attributions of high-difficulty (Hard) and low-difficulty (Easy) cohorts:
\begin{equation}
\Delta^{(i)} = \bar{s}_{hard}^{(i)} - \bar{s}_{easy}^{(i)}
\end{equation}

\begin{itemize}
    \item If $\Delta^{(i)}$ is significantly \textbf{positive}: The head is activated to identify \textbf{difficult} problems.
    \item If $\Delta^{(i)}$ is significantly \textbf{negative}: The head is activated to identify \textbf{simple} problems.
\end{itemize}

By computing $\Delta^{(i)}$ for all heads, the specific components responsible for difficulty perception can be precisely located.

\section{Additional Implementation Details}\label{Implementation_details}

\paragraph{Perplexity (PPL)}\label{PPL_details}  measures uncertainty of the generated sequence as:
\[
\text{PPL}(o^i) = \exp\left( -\frac{1}{N} \sum_{t=1}^{N} \log \pi(o_t^i \mid o_{<t}^i) \right)
\]
where $\pi(o_t^i \mid o_{<t}^i)$ is next‑token probability. Low-PPL responses are generally more fluent and semantically coherent~\cite{PPl}.

\paragraph{RL Training Configuration}\label{train_hyperparameters} For both GRPO and DAPO, we use the hyperparameters in Tab.\ref{tab_train_hyperparameters}, without using entropy or KL losses. Specifically, Branching maxtimes caps trajectory branching at 4, while Samples per prompt denotes the maximum few-shot CoT demonstrations. Experiments on Qwen2.5-7B run on \textbf{8$\times$H100 GPUs}, taking $\sim$50 hours for one epoch. Experiments on Qwen3-14B run on \textbf{16$\times$H100 GPUs}, taking $\sim$64 hours for one epoch.

\begin{table}[!htb]
\centering
\renewcommand{\arraystretch}{0.8} 
\begin{tabular}{@{}ll@{}}
\toprule
\textbf{Hyperparameter} & \textbf{Value} \\
\midrule
Optimizer & AdamW \\
Policy learning rate & $1\text{e}^{-6}$ for
qwen2\_5 series \\
& $5\text{e}^{-6}$ for qwen3 series\\
Training batch size & 128 \\
Samples per prompt & 8 \\
Rolloutn & 4 \\
Branching K & 7 \\
N & 8 \\
$\gamma$ &0.003 \\
Max response length & 8192 \\
Max prompt length & 2048 \\
Rollout temperature & 1.0 \\
Clip range $\epsilon_{\text{low}}$, $\epsilon_{\text{high}}$ & $0.2$, $0.28$ \\
$\alpha$ for IB loss & 0.005 \\
\bottomrule
\end{tabular}
\vspace{-5pt}
\caption{Hyperparameters of GRPO and DAPO training}\label{tab_train_hyperparameters}
\end{table}

\paragraph{Word Cloud}\label{world_cloud} The Entropy Reg. outputs more generic syntactic fillers (e.g., 'is', 'to', 'can'). I\textsuperscript{2}B-LPO produces more reasoning-focused keywords.

\begin{figure}[htbp]
    \centering
    \begin{subfigure}[b]{\columnwidth}
        \centering
        \includegraphics[width=0.9\linewidth]{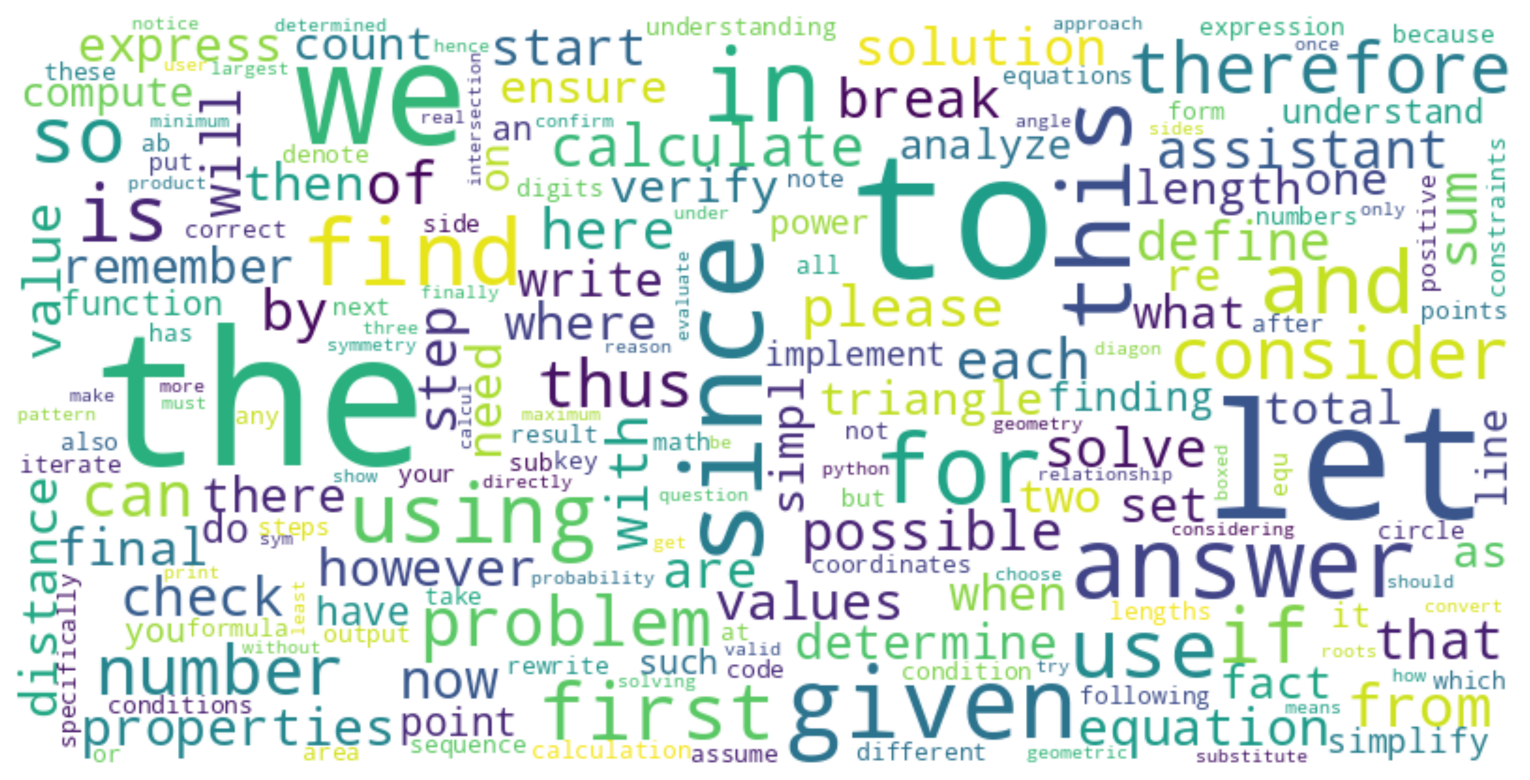}
        \caption{Naive Entropy Regularization.}
        \label{fig:wordcloud_entropy}
    \end{subfigure}
    
    \vspace{1em} 
    
    \begin{subfigure}[b]{\columnwidth}
        \centering
        \includegraphics[width=0.9\linewidth]{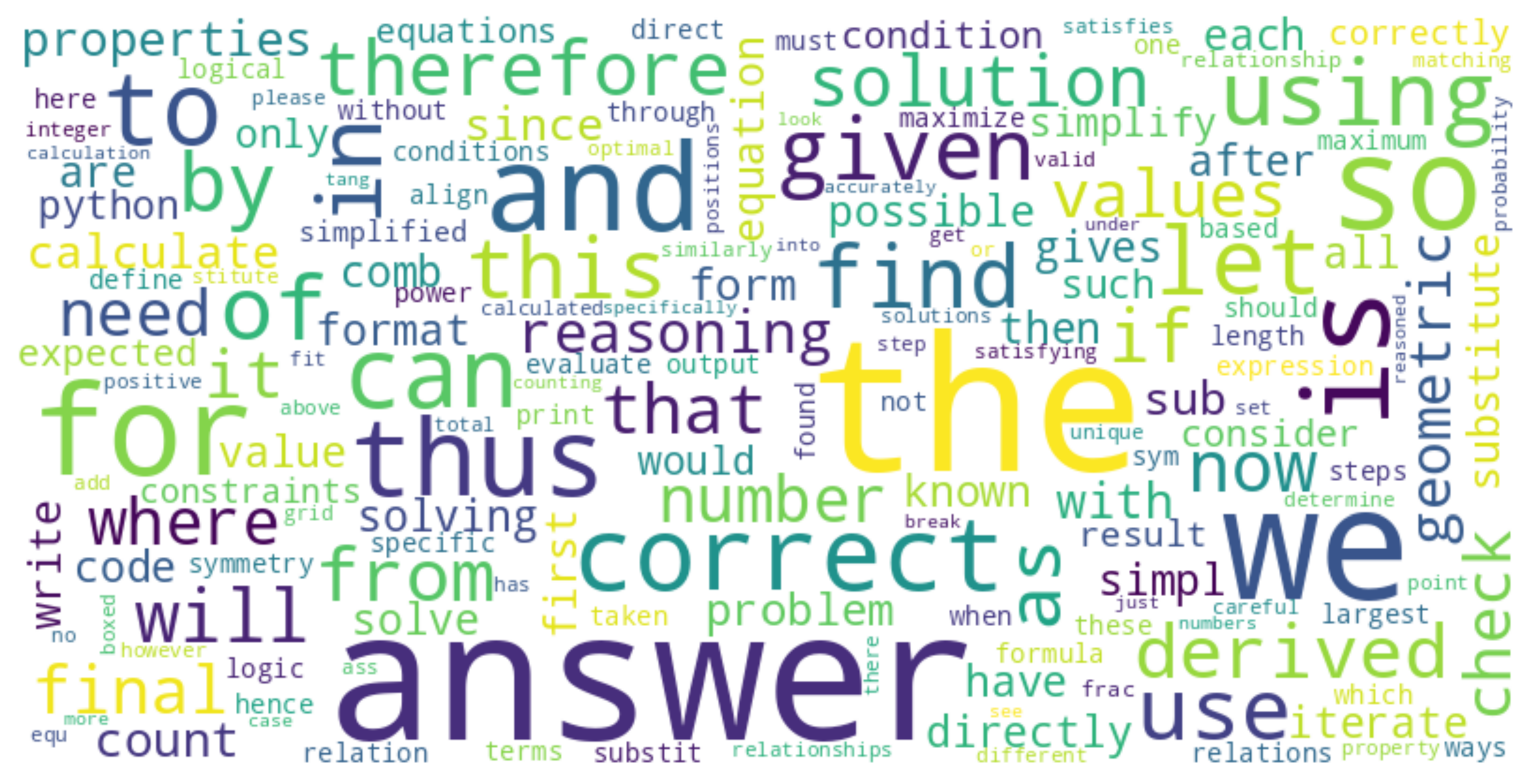}
        \caption{I\textsuperscript{2}B-LPO (Ours).}
        \label{fig:wordcloud_I2B}
    \end{subfigure}
    
    \caption{\textbf{Word cloud comparison on AIME2024 using Qwen2.5-7B. The Entropy Reg. outputs more generic syntactic fillers (e.g., 'is', 'to', 'can'). I\textsuperscript{2}B-LPO produces more reasoning-focused keywords.}}
\end{figure}

\paragraph{Response length analysis.}

I\textsuperscript{2}B-LPO achieves a balanced response length, sitting between GRPO and Entropy. Reg. It avoids the superficial verbosity and non-informative filler observed in Entropy. Reg, while demonstrating more comprehensive reasoning steps compared to GRPO.

\begin{figure}[htbp]
    \centering
    \includegraphics[width=0.95\columnwidth]{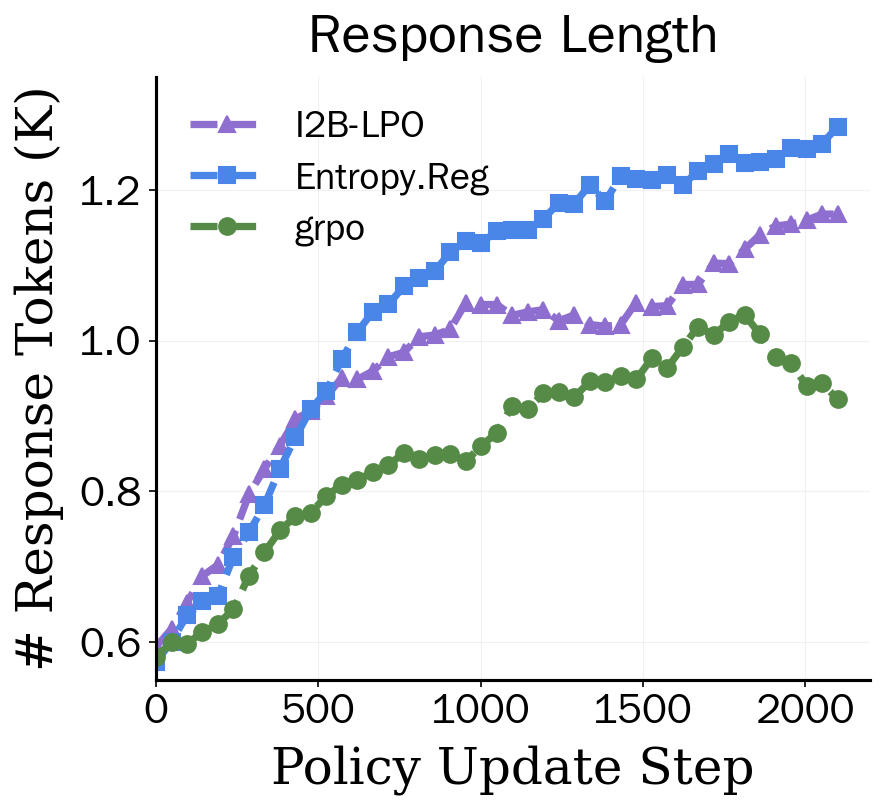}
    \vspace{-5pt}
    \caption{\textbf{Response length analysis on Qwen2.5-3B.} }
    \label{Response_length}
\end{figure}

\paragraph{Ablation of weight decay in PSA.}
\label{weight_decay}

We implement an exponential decay schedule for the injection factor $\gamma(t)$, annealing the strength from $5 \times 10^{-2}$ down to $5 \times 10^{-4}$. This experiment is conducted on MATH datasets with qwen2.5-3B base model. The Gaussian icons (Strong, Medium, Weak) visually represent the diminishing intensity of the latent variable $\mathbf{z}$ over time.

\begin{figure}[htbp]
    \centering
    \includegraphics[width=0.82\columnwidth]{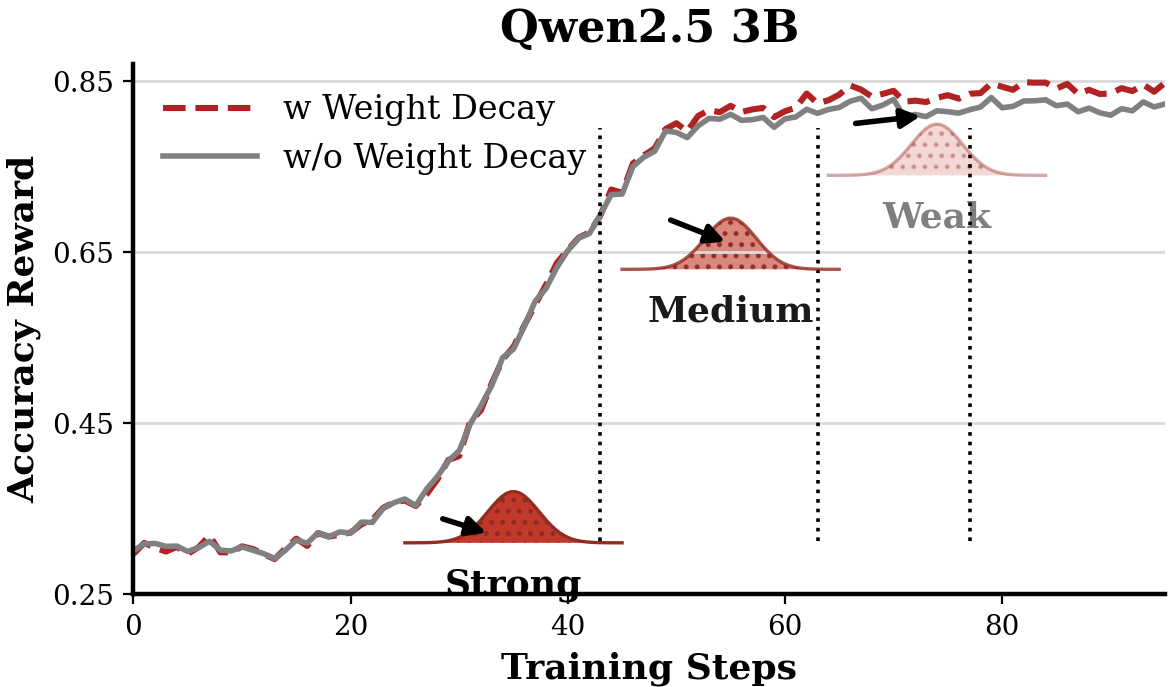}
    \vspace{-5pt}
    \caption{\textbf{Ablation study on the dynamic decay of latent injection.} }
    \label{weight_decay}
\end{figure}

\paragraph{Data Pre-processing}
\label{sec:data_filtering}

\begin{figure}[H] 
    \centering
    \vspace{-5pt} 
    
    \begin{subfigure}[b]{0.48\textwidth}
        \centering
        \includegraphics[width=\textwidth]{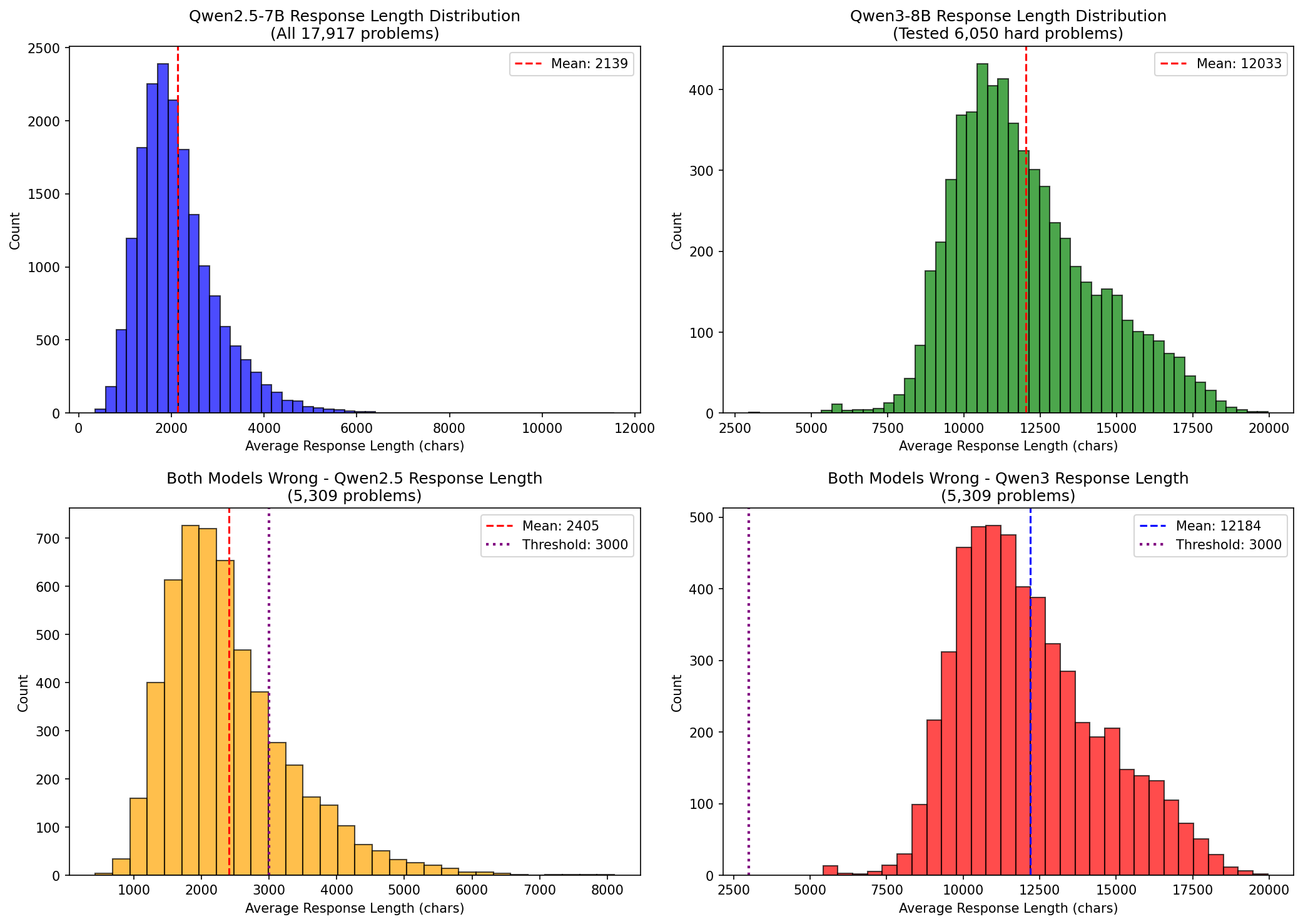} 
        \vspace{-15pt} 
        \caption{DAPO-Math-17k}
        \label{fig:dapo_dist}
    \end{subfigure}
    \hfill 
    \begin{subfigure}[b]{0.48\textwidth}
        \centering
        \includegraphics[width=\textwidth]{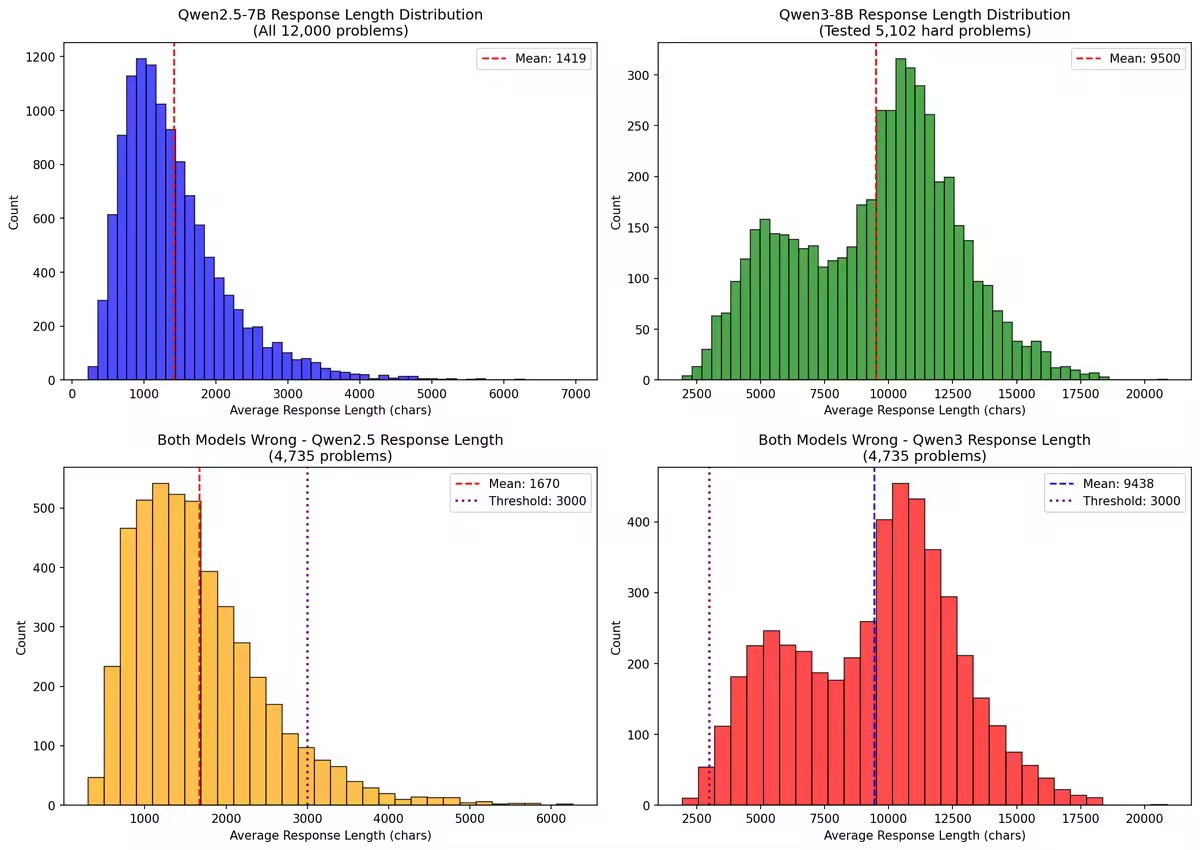}
        \vspace{-15pt} 
        \caption{MATH}
        \label{fig:math_dist}
    \end{subfigure}
    
    
    \caption{Response length distribution results on DAPO and MATH datasets.}
    \label{fig:length_distributions}
    
    \vspace{-15pt} 
\end{figure}

To focus training on the ``learning frontier", \textbf{i.e,} problems neither trivial nor intractable, we filter the dataset using two reference models (Qwen2.5-7B and Qwen3-8B) with $n=8$ rollouts each. We exclude: 
(1) \textbf{trivial samples} consistently solved by both models, as they provide negligible gradient signals for policy improvement; and 
(2) \textbf{intractable samples} where both models fail, specifically targeting excessively long responses ($>$3,000 tokens for Qwen2.5-7B, $>$10,000 for Qwen3-8B) to mitigate hallucination loops.
The response length distributions of the resulting filtered datasets are illustrated in Figure~\ref{fig:length_distributions}. The final training set consists of \textbf{6,486} samples from MATH and \textbf{13,583} from DAPO.

\end{document}